\documentclass[pdflatex,sn-mathphys-num]{sn-jnl}

\usepackage{graphicx}%
\usepackage{multirow}%
\usepackage{amsmath,amssymb,amsfonts}%
\usepackage{amsthm}%
\usepackage{mathrsfs}%
\usepackage[title]{appendix}%
\usepackage{xcolor}%
\usepackage{textcomp}%
\usepackage{manyfoot}%
\usepackage{booktabs}%
\usepackage{algorithm}%
\usepackage{algorithmicx}%
\usepackage{algpseudocode}%
\usepackage{listings}%
\usepackage{tikz}
\usepackage{pgfplots}
\usepackage{bm}

\theoremstyle{thmstyleone}%
\theoremstyle{thmstyletwo}%

\theoremstyle{thmstylethree}%

\begin{document}

\title[Article Title]{Unsupervised feature selection using Bayesian Tucker decomposition}


\author*[1]{\fnm{Y-h.} \sur{Taguchi}}\email{tag@granular.com}
\equalcont{These authors contributed equally to this work.}

\author[2]{\fnm{Yoh-ichi} \sur{Mototake}}\email{y.mototake@r.hit-u.ac.jp}
\equalcont{These authors contributed equally to this work.}

\affil*[1]{\orgdiv{Department of Physics}, \orgname{Chuo University}, \orgaddress{
\city{Tokyo}, \postcode{112-8551}, 
\country{Japan}}}

\affil[2]{\orgdiv{Graduate School of Social Data Science}, \orgname{Hitotsubashi University}, \orgaddress{
\city{Tokyo}, \postcode{186-8601}, 
\country{Japan}}}



\abstract{In this paper, we proposed Bayesian Tucker decomposition (BTuD) {\color{black} in which} residual is supposed to obey Gaussian distribution analogous to  linear regression. Although we have proposed an algorithm to perform the proposed BTuD, the conventional higher{\color{black}-}order orthogonal iteration can generate Tucker decomposition {\color{black} consistent} with the present implementation.  Using the proposed BTuD, we can perform unsupervised feature selection successfully applied to various synthetic data{\color{black}sets}, global coupled maps with randomized coupling strength, and gene expression profiles. Thus we can conclude that our newly proposed unsupervised feature selection method is promising. In addition to this, BTuD based unsupervised FE is expected to coincide with TD based unsupervised FE that were previously proposed and  successfully applied to {\color{black}a} wide {\color{black}range of} problems.}




\maketitle

\section{Introduction}\label{sec1}

The purpose of this paper is to propose a Bayesian unsupervised feature selection. There is a way to perform unsupervised feature selection using {\color{black}t}ensor decomposition (TD) by TD based unsupervised feature extraction (FE)~\cite{Taguchi2024}. In this paper, we develop Bayesian Tucker decomposition (BTuD) to perform TD based unsupervised FE in the {\color{black}framework} of Bayesian
statistics. 

TD is one of the old-fashioned embedding methods to process high dimensional data. It was invented long before the machine learning has become popular. In spite of that, it still remains one of the frequently used methods in data science. It is natural to modernize TD alongside the recent development. The {\color{black} proposal} of Bayesian TD~\cite{Cheng2023-qp} {\color{black}is} one of such {\color{black}effort}.  Nevertheless, most of those proposals are restricted to one of {\color{black}the} popular variants of TD, CP decomposition. Besides Cheng et al's book~\cite{Cheng2023-qp}, although there were many proposals of Bayesian TD~\cite{pmlr-v48-kanagawa16,NIPS2014_6d9cb7de,10.1007/s42979-022-01119-8}, most of them deal with only CP decomposition. {\color{black}Another} frequently used TD, Tucker decomposition, was  not frequently   discussed along the {\color{black}framework} of Bayes statistics~\cite{zhao2015bayesiansparsetuckermodels,10.1002/cem.1223,10305542}.   The critical difference between these a few previous implementations and the one proposed in this study is that the previous methods require that the decomposed components themselves derived from tensor decomposition obey Gaussian distribution whereas the present implementation only requires that residual{\color{black}s follow a}  Gaussian distribution.

The reason  we extend Bayesian statistic to Tucker decomposition is {\color{black}that} we would like to utilize  Bayesian TD for PCA/TD based unsupervised FE~\cite{Taguchi2024} that we have proposed {\color{black} a} long time ago. Especially, in TD based unsupervised FE, we definitely need Tucker decomposition since we have compared various TDs to establish TD based unsupervised FE and found that Tucker decomposition is the best one for TD based unsupervised FE.

The common statistical basis of the above conventional Bayesian approaches is to assume a Gaussian distribution to components themselves following the tradition of probabilistic principal component analysis (PCA)~\cite{0f9485aa-56d5-346a-95e4-aa1c6bc30fc3}, which is a bit problematic for our purposes. In PCA/TD based unsupervised feature extraction, feature selection is performed {\color{black}by} selecting outliers under the null hypothesis that components {\color{black}are} assumed to {\color{black}follow a} Gaussian distribution. Thus, since there can be no outliers when assuming that components obey Gaussian as in the other implementations of other traditional Bayesian TD, the implementation where components are assumed to be Gaussian is not adequate for our purpose. To address this problem, instead of assuming the components themselves as Gaussian, employing the framework of linear regression, we assume that not the components themselves but the residuals obey Gaussian. Since the implementation of linear regression using Bayes statistics is well known~\cite{M2006-ab}, to get a representation of BTuD within this strategy is straightforward.


\section{Previous feature selection approaches}

Most previous feature selection approaches are supervised ones. There are two branches of supervised feature selection.
One is {\color{black}the} filter {\color{black}approach} and {\color{black}the other} is {\color{black}the wrapper approach}. In the {\color{black}filter} approaches, individual features are evaluated based upon provided labels/classification. One of typical filter approaches
is \textit{t} test. In the \textit{t}{\color{black}-}test, $P$-values are attribute{\color{black}d} to individual features with assuming null hypothesis that {\color{black}all} components of individual features are drawn from {\color{black}the same} distribution and features that can reject the null hypothesis are selected. One of {\color{black}wrapper} approaches is random forest~\cite{ho1995random}. In random forest, 
individual features are randomly selected and replace{\color{black}d} until the {\color{black}desired} performance is achieved. Anyway, in the supervised approaches, we need some (external) supervised criteria by which we can evaluate individual features.

Unsupervised approach is completely different from the supervised approaches. In unsupervised approaches, we have to have criteria by which we can evaluate individual features without any external information. One typical unsupervised approache is unimodality testing~\cite{fb099965-2cf5-3955-b5e5-077ac9290f24}. In unimodality testing, components in {\color{black}an} individual feature {\color{black}are} assumed to be unimodal (i.e., {\color{black}the} distribution do{\color{black}es} not have multiple peaks but have only one peak).    Although unimodality testing is powerful, it has limited ability to select features since features associated with {\color{black} single-peaked} distribution  cannot be selected even if it has some order. Thus, we need more powerful unsupervised approaches.
PCA/TD based unsupervised FE is such a method, since the {\color{black}criterion for} selecting features is independent of the evaluation of features (see Methods).
Once singular value vectors are computed with attributing some of them to samples and others to features. Evaluation is based upon those attributed to samples and selection is based upon those attributed to features. Since the evaluation and selection is separated, we can select any kind of features coincident with external labels. 
Thus to perform TD based unsupervised FE in Bayesian statistics is promising approach and {\color{black}worthwhile} trying.

\section{Methods}\label{sec11}

\subsection{Implementation of Bayesian Tucker decomposition with a framework of linear regression}

Assume that we have Tucker decomposition of {\color{black}third-order} tensor, $x_{ijk} \in \mathbb{R}^{N \times M \times K}$
as
\begin{equation}
x_{ijk} \simeq \sum_{\ell_1=1}^{L_1} \sum_{\ell_2=1}^{L_2} \sum_{\ell_3=1}^{L_3}
G(\ell_1 \ell_2 \ell_3)  u_{\ell_1 i}u_{\ell_2 j}u_{\ell_3 k}  \label{eq:Tucker}
\end{equation}
where $1 \leq L_1 \leq N, 1\leq L_2 \leq M, 1 \leq  L_3 \leq K$ and $G \in \mathbb{R}^{L_1 \times L_2 \times L_3}$ is a core tensor that represents the weight of the product $u_{\ell_1 i}u_{\ell_2 j}u_{\ell_3 k}$ toward $x_{ijk}$, $u_{\ell_1 i} \in \mathbb{R}^{L_1 \times N} $, $u_{\ell_2 j} \in \mathbb{R}^{L_2 \times M} $, and $u_{\ell_3 k} \in \mathbb{R}^{L_3 \times K}$ are singular value matrices and orthogonal matrices.

\begin{figure}
    \centering
    \includegraphics[width=\linewidth]{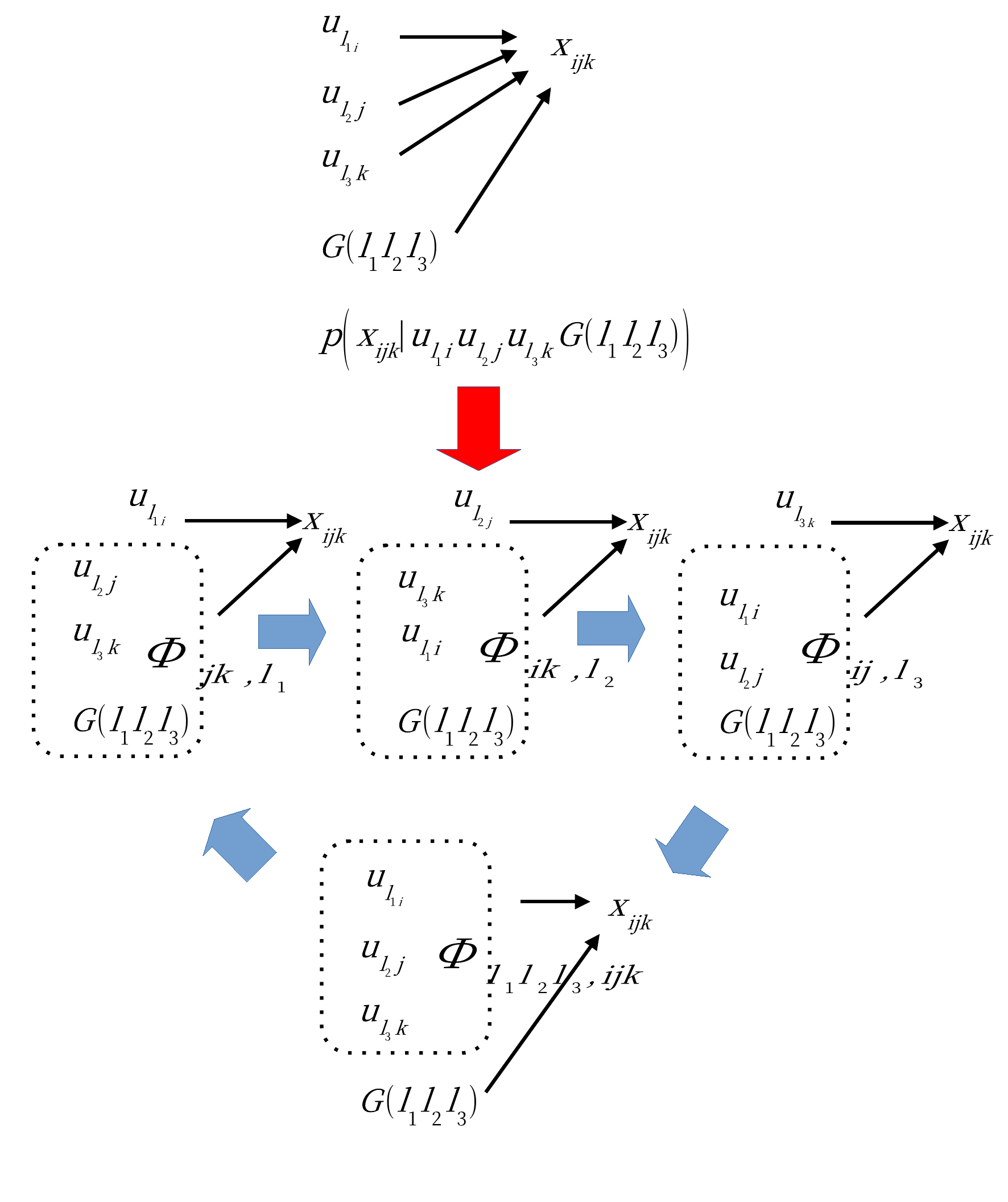}
    \caption{ Graph{\color{black}ical} diagram that shows the structure of Bayesian inference. Top row is consistent with  Eq. (\ref{eq:Tucker}). 
    The middle row is consistent with eqs. (\ref{eq1}), (\ref{eq2}), and (\ref{eq3}) (From left to right panels in the middle row).  The bottom row is consistent with eq. (\ref{eq4}).
    The cyclic  computation among  the middle and bottom rows is the approximation to the first row. } .
    \label{fig:graph_diagram}
\end{figure}

Top row of Fig. \ref{fig:graph_diagram} shows the graph diagram to show the structure of Bayesian inference consistent with eq. (\ref{eq:Tucker}).
The likelihood that corresponds to the top row of Fig. \ref{fig:graph_diagram} (or eq. (\ref{eq:Tucker})) is written as 
\begin{equation}
p(x_{ijk} | u_{\ell_1 i}, u_{\ell_2 j}, u_{\ell_3 k},  G(\ell_1 \ell_2\ell_3) ) \label{eq:likelihood}
\end{equation}
and prior is written as
\begin{equation}
    p(u_{\ell_1 i}, u_{\ell_2 j}, u_{\ell_3 k},  G(\ell_1 \ell_2\ell_3) ) \label{eq:prior}
\end{equation}
Thus the posterior is
\begin{equation}
    p( u_{\ell_1 i}, u_{\ell_2 j}, u_{\ell_3 k},  G(\ell_1 \ell_2\ell_3) | x_{ijk} ) = \frac{ p(x_{ijk} | u_{\ell_1 i}, u_{\ell_2 j}, u_{\ell_3 k},  G(\ell_1 \ell_2\ell_3) )  p(u_{\ell_1 i}, u_{\ell_2 j}, u_{\ell_3 k},  G(\ell_1 \ell_2\ell_3) ) }{p(x_{ijk})} \label{eq:postrior}
\end{equation}

\subsubsection{Bayesian Tucker decomposition interpreted as a set of linear regression}
Unfortunately, since it is not very practical to infer  {\color{black}Eq.} (\ref{eq:Tucker}) directly, {\color{black}so} we {\color{black}decompose}  it to a set of linear regression{\color{black}s} as follows.

To apply Bayesian linear regression to Tucker decomposition we re-write it as
\begin{eqnarray}
     \Phi_{jk,\ell_1} & \equiv& \sum_{\ell_2,\ell_3} G(\ell_1 \ell_2 \ell_3 ) u_{\ell_2 j} u_{\ell_3 k} \in \mathbb{R}^{MK \times L_1}\\
    x_{ijk} &  \simeq& \sum_{\ell_1}   \Phi_{jk,\ell_1} u_{\ell_1 i} \label{eq1}\\
     \Phi_{ik,\ell_2} & \equiv & \sum_{\ell_1,\ell_3} G(\ell_1 \ell_2 \ell_3 ) u_{\ell_1 i} u_{\ell_3 k} \in \mathbb{R}^{NK \times L_2}\\
     x_{ijk} &  \simeq& \sum_{\ell_2}   \Phi_{ik,\ell_2} u_{\ell_2 j} \label{eq2} \\
      \Phi_{ij,\ell_3} & \equiv & \sum_{\ell_1,\ell_2} G(\ell_1 \ell_2 \ell_3 ) u_{\ell_1 i} u_{\ell_2 j} \in \mathbb{R}^{ NM \times L_3}\\
      x_{ijk} &  \simeq& \sum_{\ell_3}   \Phi_{ij,\ell_3} u_{\ell_3 k} \label{eq3}\\
      \Phi_{\ell_1 \ell_2 \ell_3, i j k} & \equiv &
      u_{\ell_1 i} u_{\ell_2 j} u_{\ell_3 k} \in \mathbb{R}^{L_1 L_2 L_3 \times NMK} \\
      x_{ijk} &  \simeq & \sum_{\ell_1 \ell_2 \ell_3} \Phi_{\ell_1 \ell_2 \ell_3, i j k} G(\ell_1 \ell_2 \ell_3) \label{eq4}
\end{eqnarray}
{\color{black}where E}qs. (\ref{eq1}), (\ref{eq2}), (\ref{eq3}), and (\ref{eq4}) are regarded as linear regression problems where $u_{\ell_1 i}$, 
$u_{\ell_2 j}$, $ u_{\ell_3 k}$, and $G(\ell_1 \ell_2 \ell_3)$ are the regression coefficients with the given $\Phi_{jk,\ell_1}$,
$\Phi_{ik,\ell_2}$, $ \Phi_{ij,\ell_3}$, and $ \Phi_{\ell_1 \ell_2 \ell_3, i j k} $.

Actual procedures are as follows. When applying linear regression to eq. (\ref{eq1}), at first, $i$ is fixed, $u_{\ell_1 i}$ is assumed to be the {\color{black}set of} regression coefficients{\color{black}, with} $L_1$ {\color{black}coefficients}, and $\Phi_{jk,\l_1}$ is supposed to be given (i.e., $u_{\ell_2 j}$  and
$u_{\ell_3 k}$ are supposed to be known). Thus, there are $L_1$ variables whereas the number of equations (conditions) is $MK$. This means that 
$L_1$ can be larger than $MK$. To avoid {\color{black}the} difficulty {\color{black}in} this case, we employed Moore-Penrose pseudoinverse~\cite{10.1007/s13538-011-0052-z} matrix to perform linear regression (see below). 

This is also repeated toward $u_{\ell_2 j}$ (eq. (\ref{eq2}) or the middle panel of the middle row of Fig. \ref{fig:graph_diagram}) and $u_{\ell_3 k}$ (eq. (\ref{eq3}) or the right panel of the middle row of Fig. \ref{fig:graph_diagram}) with
cycling indices among $i,j,k$ and $\ell_1,\ell_2,\ell_3$. The graphical diagram of this procedure is in the middle  of Fig. \ref{fig:graph_diagram}.
After that, the linear regression that corresponds to  eq. (\ref{eq4}) (or the bottom row of Fig. \ref{fig:graph_diagram}) is performed. 
The whole process that includes these four linear regression is an approximation to eq. (\ref{eq:Tucker}) (or top row of Fig. \ref{fig:graph_diagram})

This means that
\begin{eqnarray}
{\color{black}
u_{\ell_a i_a}^{\mbox{\tiny MAP}} = \arg\max_{u_{\ell_a i_a}} p( u_{\ell_a i_a}, u^{\mbox{MAP}}_{\ell_b i_b}, u^{\mbox{MAP}}_{\ell_c i_c},  G^{\mbox{MAP}}(\ell_1 \ell_2 \ell_3) | x_{ijk} )} \\\label{eq:cyc} 
G^{\mbox{MAP}}(\ell_1 \ell_2\ell_3) = \arg\max_{G(\ell_1 \ell_2\ell_3) } p( u^{\mbox{MAP}}_{\ell_1 i}, u^{\mbox{MAP}}_{\ell_2 j}, u^{\mbox{MAP}}_{\ell_3 k},  G(\ell_1 \ell_2\ell_3) | x_{ijk} )  
\end{eqnarray}
{\color{black} where $(\ell_a, \ell_b, \ell_c) =( \ell_1,\ell_2, \ell_3)$ and $(i_a, i_b, i_c) =( i, j, k)$ in eq. (\ref{eq:cyc}) should be cyclically changed.}

Although likelihood, eq. (\ref{eq:likelihood}), does not change, prior, Eq. (\ref{eq:prior}), that corresponds to 
  {\color{black} eqs. (\ref{eq1}), (\ref{eq2}), and (\ref{eq3}) (or the left to right  panels of the middle row of Fig. \ref{fig:graph_diagram})  are re-written as}
\begin{eqnarray}
  && p(u_{\ell_1 i}, u_{\ell_2 j}, u_{\ell_3 k},  G(\ell_1 \ell_2\ell_3) ) \nonumber \\ & = & p(u_{\ell_1 i})p(u_{\ell_2 j}, u_{\ell_3 k},  G(\ell_1 \ell_2\ell_3)) \nonumber\\
   &= &p(u_{\ell_a i_a}) \delta(u_{\ell_b i_b} -u_{\ell_b i_b}^{\mbox{\tiny MAP}})
    \delta(u_{\ell_c i_c} -u_{\ell_c i_c}^{\mbox{\tiny MAP}})  \delta(G(\ell_1 \ell_2 \ell_3) -G^{\mbox{\tiny MAP}}(\ell_1 \ell_2 \ell_3)) \;
\end{eqnarray}
{\color{black} where $(\ell_a, \ell_b, \ell_c) =( \ell_1,\ell_2, \ell_3)$ and $(i_a, i_b, i_c) =( i, j, k)$ should be cyclically changed as well.
and $p(u_{\ell_a i_a})$, $p(u_{\ell_b i_b})$, and $p(u_{\ell_c i_c})$ are the uniform distribution{\color{black}s}.}
Thus, posterior, eqs. (\ref{eq:postrior}),  also re-written as
\begin{eqnarray}
   && p( u_{\ell_1 i}, u_{\ell_2 j}, u_{\ell_3 k},  G(\ell_1 \ell_2\ell_3) | x_{ijk} )  \nonumber \\
    &= &  \frac{p(x_{ijk} | u_{\ell_1 i}, u_{\ell_2 j}, u_{\ell_3 k},  G(\ell_1 \ell_2\ell_3) )  p(u_{\ell_1 i}, u_{\ell_2 j}, u_{\ell_3 k},  G(\ell_1 \ell_2\ell_3) ) }{p(x_{ijk})} \nonumber \\
     & =& {\color{black}\frac{p(x_{ijk} | u_{\ell_a i_a}, u_{\ell_b i_b}^{\mbox{\tiny MAP}}, u_{\ell_c i_c}^{\mbox{\tiny MAP}},  G^{\mbox{\tiny MAP}}(\ell_1 \ell_2\ell_3) )  p(u_{\ell_a i_a}) }{p(x_{ijk})} }
\end{eqnarray}

\begin{figure}
    \centering
    \includegraphics[width=\linewidth]{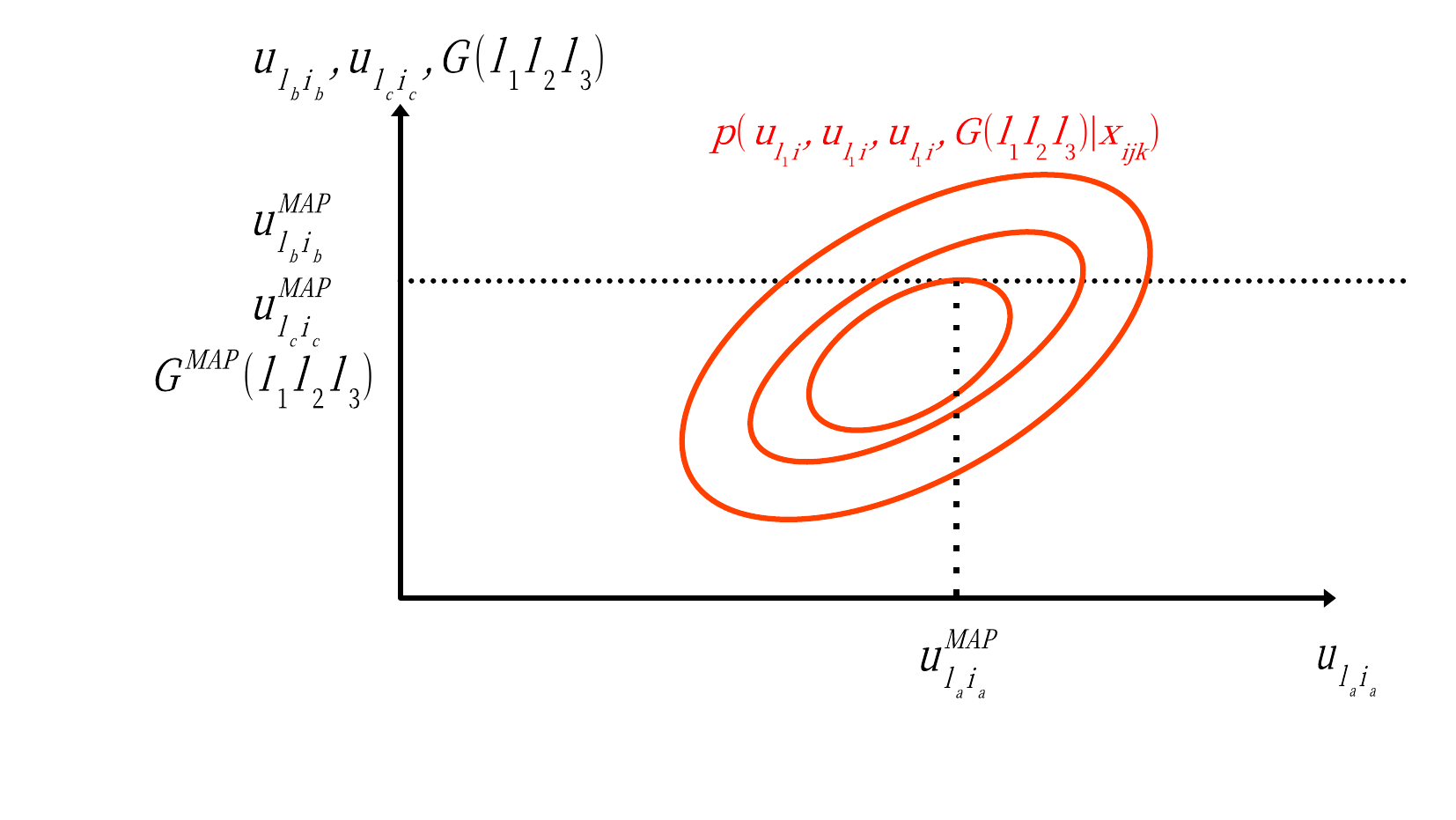}
    \caption{ The process that optimizes posterior, $p( u_{\ell_1 i}, u_{\ell_2 j}, u_{\ell_3 k},  G(\ell_1 \ell_2\ell_3) | x_{ijk} )$, which is represented by red contour, towards {\color{black} $u_{\ell_a i_a}$ with fixed values $ u_{\ell_b i_b}^{\mbox{\tiny MAP}}, u_{\ell_c i_c}^{\mbox{\tiny MAP}},  G^{\mbox{\tiny MAP}}(\ell_1 \ell_2\ell_3)$ where $(\ell_a, \ell_b, \ell_c) =( \ell_1,\ell_2, \ell_3)$ and $(i_a, i_b, i_c) =( i, j, k)$  should be cyclically changed.} This does not clearly give us global minimum but local minimum. }
    \label{fig:local_minimum}
\end{figure}

The optimization process for {\color{black} eqs. (\ref{eq1}), (\ref{eq2}), and (\ref{eq3}) (or the left to right panels} of the middle row of Fig. \ref{fig:graph_diagram})  is illustrated in Fig. \ref{fig:local_minimum}.

{\color{black} In summary, we decomposed Tucker decomposition eq. (\ref{eq:Tucker}) into four linear regression subproblems, each of which  estimates one of $u_{\ell_1 i}$, $u_{\ell_2 j}$, $u_{\ell_3 k}$, and $G(\ell_1 \ell_2 \ell_3)$ as MAP solution.
In each subproblem, the remaining three variables  are treated as  constants fixed at their current MAP estimates.
Since this alternating optimization updates one subproblem at a time,  it is not guaranteed to reach the global minimum and may instead converge to a local minimum  (Fig. \ref{fig:local_minimum}). Nevertheless,  the obtained solution is expected to be close to the global minimum unless the optimization is trapped in a poor local minimum.  
}

\subsection{Likelihood}

Let $\mathbf{x}_i\in\mathbb{R}^{MK}$ be the vectorization of $\{x_{ijk}\}_{j,k}$,
and let $\mathbf{\Phi}\in\mathbb{R}^{MK\times L_1}$ be the design matrix with
$\Phi_{(j,k),\ell_1}=\Phi_{jk,\ell_1}$ in Eq.(2). Then Eq.(3) is written as
\[
\mathbf{x}_i = \mathbf{\Phi}\,\mathbf{u}_i + \boldsymbol{\varepsilon}_i,
\qquad
\boldsymbol{\varepsilon}_i \sim \mathcal{N}(\mathbf{0},\beta^{-1}\mathbf{I}),
\]
which yields the likelihood
\[
p(\mathbf{x}_i \mid \mathbf{u}_i,\mathbf{\Phi},\beta)
= \mathcal{N}\!\left(\mathbf{x}_i \mid \mathbf{\Phi}\mathbf{u}_i,\beta^{-1}\mathbf{I}\right).
\]
Equivalently, for the whole tensor,
\[
p(\{x_{ijk}\}\mid \{u_{\ell_1 i}\},\{u_{\ell_2 j}\},\{u_{\ell_3 k}\},G,\beta)
= \prod_{i,j,k}\mathcal{N}\!\Bigl(x_{ijk}\mid
\sum_{\ell_1,\ell_2,\ell_3}G_{\ell_1\ell_2\ell_3}u_{\ell_1 i}u_{\ell_2 j}u_{\ell_3 k},
\beta^{-1}\Bigr).
\]

     \subsection{Prior probability} 

     We place independent zero-mean Gaussian priors on the regression coefficients:
\[
p(\mathbf{u}_i\mid \alpha)=\mathcal{N}(\mathbf{u}_i\mid \mathbf{0},\alpha^{-1}\mathbf{I}),
\quad
p(\mathbf{u}^{(2)}_j\mid \alpha)=\mathcal{N}(\mathbf{u}^{(2)}_j\mid \mathbf{0},\alpha^{-1}\mathbf{I}),
\quad
p(\mathbf{u}^{(3)}_k\mid \alpha)=\mathcal{N}(\mathbf{u}^{(3)}_k\mid \mathbf{0},\alpha^{-1}\mathbf{I}),
\]
and similarly for the vectorized core tensor $\mathbf{g}=\mathrm{vec}(G)$:
\[
p(\mathbf{g}\mid \alpha)=\mathcal{N}(\mathbf{g}\mid \mathbf{0},\alpha^{-1}\mathbf{I}).
\]



\subsection{Posterior Probability}

{\color{black}The p}osterior probability of $u_{\ell_1 i}, {\color{black}u_{\ell_2 j}}, u_{\ell_3 k}, G(\ell_1, \ell_2, \ell_3)$ are given as follows. 

{\color{black}
\begin{equation}
    P\left (\{u_{\ell_a i_a}\}_{\ell_a} | x_{ijk}, \{G\}_{\ell_b,\ell_c}, 
    \{m_{u_{\ell_b i_b}}\}_{\ell_b}, \{ m_{u_{\ell_c i_c}} \}_{\ell_c} \right) =  \mathcal{N} \left ( \{ u_{\ell_a i_a} \}_{\ell_a} |\{ m_{u_{\ell_a i_a}} \}_{\ell_a},
    \{ S_{\ell_a \ell_a'} \}_{\ell_a,\ell_a'} \right ) \label{eq:post1}
    \end{equation}
    \begin{equation}
    m_{u_{\ell_a i_a}}   = \left \lbrace \begin{array}{lc}  \sum_{i_b i_c} \Phi_{i_b i_c,\ell_a}^\dagger x_{ijk}   &(\alpha=0)\\
                 \sum_{i_b i_c\ell_a'}  \beta   S_{\ell_a \ell_a'}   \Phi_{i_b i_c,\ell_a'}   x_{ijk}   & (\alpha \neq 0)
                          \end{array} \right . \in \mathbb{R}^{L_1 \times N} \label{eq:m}
          \end{equation} 
          \begin{equation}
    S_{\ell_a \ell_a'}^{-1}  =   \alpha I+ \beta \sum_{i_b i_c}  \Phi_{i_b i_c,\ell_a} \Phi_{i_b i_c,\ell_a'} \in \mathbb{R}^{L_1 \times L_1}
    \end{equation}
    
    where $(\ell_a, \ell_b, \ell_c) =( \ell_1,\ell_2, \ell_3)$ and $(i_a, i_b, i_c) =( i, j, k)$ should be cyclically changed.
    }

     \begin{equation}
    P\left (\{ G \}_{\bm{\ell}}  | x_{ijk} , \{m_{u_{\ell_1 i}}\}_{\ell_1}, \{m_{u_{\ell_2 j}}\}_{\ell_2}  , \{m_{u_{\ell_3 k}}\}_{\ell_3} \right ) =  \mathcal{N}\left( \{ G \}_{\bm{\ell}} | \{ m_{G} \}_{\bm{\ell}}, \{ S_{\bm{\ell} \bm{\ell}' } \}_{\bm{\ell} \bm{\ell}' } \right) 
    \label{eq:post4}
     \end{equation}
    \begin{equation}
    m_{G}  =  \left \lbrace \begin{array}{lc}
    \sum_{ijk}\Phi_{\ell_1 \ell_2 \ell_3, i  j k}^\dagger x_{ijk}     & (\alpha=0)  \\
    \sum_{ijk \bm{\ell}'} \beta S_{\bm{\ell} \bm{\ell}'}  \Phi_{\bm{\ell}, i  j k} x_{ijk}   & (\alpha \neq 0) \end{array}
   \right .  \in \mathbb{R}^{L_1 L_2 L_3 \times L_1 L_2 L_3} \end{equation}
    \begin{equation}
     \end{equation}
    \begin{equation}
    S_{\bm{\ell} \bm{\ell}'}^{-1}  =   \alpha I+\beta \sum_{ijk}  \Phi_{\bm{\ell}, i  j k}  \Phi_{\bm{\ell}', i  j k}   \in \mathbb{R}^{L_1 L_2 L_3 \times L_1 L_2 L_3}
     \end{equation}
where, for example, $\{ \cdots \}_{\ell_1}$ means that it is composed of
$1 \leq \ell_1 \leq L_1$ and so on whereas $\bm{\ell} = (\ell_1, \ell_2, \ell_3)$ and $\bm{\ell}' = (\ell_1', \ell_2', \ell_3')$.

More explicitly,
{\color{black}
\begin{equation}
    P\left (\{u_{\ell_a i_a}\}_{\ell_a} | x_{ijk}, \{ G \}_{\ell_b,\ell_c}, \{m_{u_{\ell_b i_b}}\}_{\ell_b}, \{m_{u_{\ell_c i_c}} \}_{\ell_c} \right)
     =  \frac{e ^{ -\frac{1}{2}  \left \lbrace \sum_{\ell_a,\ell_a'=1}^{L_1} \left (u_{\ell_a i_a} -  m_{u_{\ell_a i_a}} \right)
      \left [ S_{\ell_a, \ell_a'}^{-1} \right ]_{\ell_a \ell_a'}
      \left (u_{\ell_a' i_a} -  m_{u_{\ell_a' i_a}}\right)\right \rbrace} }{\sqrt{ (2\pi)^{L_a}\left | S_{\ell_a,\ell_a'} \right |}} 
      \end{equation}
      where $(\ell_a, \ell_b, \ell_c) =( \ell_1,\ell_2, \ell_3)$ and $(i_a, i_b, i_c) =( i, j, k)$ should be cyclically changed and $L_a$ should also be changed as $(L_1,L_2,L_3)$ accordingly.
      }

\subsection{Algorithm}

As mentioned above, $u_{\ell_1 i}$, $u_{\ell_2 j}$, and $u_{\ell_3 k}$ cannot be fitted simultaneously but {\color{black}must be fitted one by one}. To do this, we proposed the following procedure (Algorithm \ref{algorithm}) where $A^\dagger = (A^T A)^{-1} A^T$is Moore–Penrose inverse. $\alpha$ is the multiplier of $l_2$-norm (do not confuse $l_2$ of $l_2$-norm with index $\ell_2$ in $G (\ell_1 \ell_2 \ell_3)$ or $u_{\ell_2 j}$).

The outline of Algorithm \ref{algorithm} is as follows. At first, $u_{\ell_1 i}$ is tried to be computed with eq. (\ref{eq1}).
Orthogonalization as well as normalization are applied to $u_{\ell_1 i}$.
$G$ is tried to be computed  with eq. (\ref{eq4}). Then, $u_{\ell_2 j}$ is tried to be computed with eq. (\ref{eq2}). Orthogonalization as well as normalization are applied to $u_{\ell_2 j}$. $G$ is tried to be computed  with eq. (\ref{eq4}). Finally, $u_{\ell_3 k}$ is tried to be computed with eq. (\ref{eq3}). Orthogonalization as well as normalization are applied to $u_{\ell_3 k}$.  $G$ is tried to be computed  with eq. (\ref{eq4}). These whole processes are repeated until all of $u_{\ell_1 i}$, $u_{\ell_2 j}$, $u_{\ell_3 k}$, and $G$ are converged. 

The critical difference  between HOOI and the Algorithm \ref{algorithm} is that the former employs SVD whereas the latter employs {\color{black}linear} regression. In the sense that they both employ the scheme of alternative least squares, they are the same as others. 
\begin{algorithm}
\caption{Bayesian Tucker decomposition with linear regression}\label{algorithm}
\begin{algorithmic}[1]
\Require  $G, u_{\ell_1 i}, u_{\ell_2 j}, u_{\ell_3 k}$ computed by HOOI. 
\While{not converged}
    \While{$\ell_1 \leq L_1$} \state{ [Solve eq. (\ref{eq1}) ]}

      \state $\Phi_{jk,l_1}  \gets \sum_{\ell_2,\ell_3} G(\ell_1 \ell_2 \ell_3 ) u_{\ell_2 j} u_{\ell_3 k} $

        \While{$i \leq N$}

\If {($\alpha=0$)}
\state  $u_{\ell_1 i} \gets  \sum_{jk} \Phi_{jk,\ell_1}^\dagger x_{ijk}$ 
\Else 
\state { $u_{\ell_1 i}  \gets \sum_{\ell_1'}   \left[ \left(\sum_{jk} \Phi_{jk,\ell_1}\Phi_{jk,\ell_1'} + \alpha I \right)^\dagger \right]_{\ell_1 \ell_1'} \sum_{jk} \Phi_{jk,\ell_1'} x_{ijk}  $ } 
\EndIf

 \state{$i \gets i+1$}

 \EndWhile

\state  $u_{\ell_1 i} \gets u_{\ell_1 i} \perp  u_{\ell_1' \neq \ell_1 i}$ \state{ [Orthogonalization] }

\state $u_{\ell_1 i} \gets u_{\ell_1 i} / \sqrt{\sum_i u_{\ell_1 i}^2}$
\state{[Normalization]}

\state $\Phi_{\ell_1 \ell_2 \ell_3, i j k} \gets
      u_{\ell_1 i} u_{\ell_2 j} u_{\ell_3 k}$
      
\state $G(\ell_1 \ell_2 \ell_3 ) \gets 
\sum_{ijk} \Phi_{\ell_1 \ell_2 \ell_3, i  j k}^\dagger x_{ijk} $ \state{[Solve Eq. (\ref{eq4})]}

  \state{$\ell_1 \gets \ell_1+1$}
 
 \EndWhile
 
\state {  }

\While{$\ell_2 \leq L_2$}   \state{ [Solve eq. (\ref{eq2}) ]}

\state $\Phi_{ik,l_2} \gets \sum_{\ell_1,\ell_3} G(\ell_1 \ell_2 \ell_3 ) u_{\ell_1 i} u_{\ell_3 k}$

\While{$j \leq M$} 

\If{($\alpha=0$)}
\state
$u_{\ell_2 j}  \gets \sum_{ik} \Phi_{ik,\ell_2}^\dagger x_{ijk} $
\Else
\state {  $u_{\ell_2 j}  \gets \sum_{\ell_2'}  \left[  \left(\sum_{ik} \Phi_{ik,\ell_2}\Phi_{ik,\ell_2'} + \alpha I \right)^\dagger   \right]_{\ell_2 \ell_2'}\sum_{ik} \Phi_{ik,\ell_2'} x_{ijk}  $ }
\EndIf 
\state $j \gets j+1$
\EndWhile

\state $u_{\ell_2 j} \gets u_{\ell_2 j} \perp  u_{\ell_2' \neq \ell_2 j}   $ \state{[Orthogonalization]} 

\state $u_{\ell_2 j} \gets u_{\ell_2 j} / \sqrt{\sum_j u_{\ell_2 j}^2}$
\state{[Normalization]} 

\state $\Phi_{\ell_1 \ell_2 \ell_3, i j k} \gets
      u_{\ell_1 i} u_{\ell_2 j} u_{\ell_3 k}$
      
\state $ G(\ell_1 \ell_2 \ell_3 ) \gets
\sum_{ijk} \Phi_{\ell_1 \ell_2 \ell_3, i  j k}^\dagger x_{ijk} $ \state{[Solve Eq. (\ref{eq4})]}

\state $\ell_2 \gets \ell_2+1$
\EndWhile

\While{$\ell_3 \leq L_3$} \state{[Solve Eq. (\ref{eq3})]}

\state $ \Phi_{ij,l_3}\gets\sum_{\ell_1,\ell_2} G(\ell_1 \ell_2 \ell_3 ) u_{\ell_1 i} u_{\ell_2 j}  $

\While{$k \leq K$}

\If{($\alpha=0$)}
\state
$u_{\ell_3 k}  \gets  \sum_{ij} \Phi_{ij,\ell_3}^\dagger x_{ijk} $
\Else
{ $u_{\ell_3 k}  \gets \sum_{\ell_3'}  \left[\left(\sum_{ij} \Phi_{ij,\ell_3}\Phi_{ij,\ell_3'} + \alpha I \right)^\dagger \right]_{\ell_3 \ell_3'} \sum_{ij} \Phi_{ij,\ell_3'} x_{ijk}  $}
\EndIf

\state $k \gets k+1$
\EndWhile

\state $u_{\ell_3 k} \gets u_{\ell_3 k} \perp  u_{\ell_3' \neq \ell_3 k} $ \state{[Orthogonalization]} 

\state $u_{\ell_3 k} \gets u_{\ell_3 k} / \sqrt{\sum_k u_{\ell_3 k}^2}$
\state{[Normalization]} 

\state $\Phi_{\ell_1 \ell_2 \ell_3, i j k} \gets
      u_{\ell_1 i} u_{\ell_2 j} u_{\ell_3 k}$

\state $G(\ell_1 \ell_2 \ell_3 ) \gets
\sum_{ijk} \Phi_{\ell_1 \ell_2 \ell_3, i  j k}^\dagger x_{ijk} $  \state{[Solve Eq. (\ref{eq4})]}

$\ell_3 \gets \ell_3+1$
\EndWhile
\EndWhile
\end{algorithmic}
\end{algorithm}

\subsection{Replacement with HOOI}

Although we have proposed the algorithm, Algorithm \ref{algorithm}, to perform BTuD, we occasionally found that HOOI can also give us the solution coincident with the proposed BTuD when $\alpha=0$.
Empirically, if $u_{\ell_1i}$ is equal to  $m_{u_{\ell_1 i}}$, $u_{\ell_2j}$ is equal to  $m_{u_{\ell_2 j}}$, $u_{\ell_3k}$ is equal to  $m_{u_{\ell_3 k}}$, and
$G(\ell_1 \ell_2 \ell_3)$ is equal to  $m_G$, they can be regarded as solutions for BTuD.
Thus, when the solution computed by HOOI is coincident with $m_{u_{\ell_1 i}}, m_{u_{\ell_2 j}}$, $m_{u_{\ell_3 k}}$ and $m_G$, we do not execute Algorithm \ref{algorithm}, but employ the solution by HOOI as it is.
This means that, with starting initial values computed by HOOI, Algorithm 1 converges immediately. Also one should notice that this does not always mean the solution given by HOOI is identical to that given by Algorithm \ref{algorithm}, since there are no uniqueness about the solutions coincident with the proposed BTM. In this study, we always employed the solution by HOOI, since we must anyway execute HOOI at the very first stage in Algorithm \ref{algorithm} and always found that it is the solution coincident with $m_{u_{\ell_1 i}}, m_{u_{\ell_2 j}}$, $m_{u_{\ell_3 k}}$ and $m_G$ (in this study, we considered only $\alpha=0$). 
For HOOI, we employed \verb+tucker+ function in rTensor~\cite{rTensor} package in R.

\subsection{TD based unsupervised FE}

Before explaining how we can make use of  the above BTuD for feature selection, we introduce the previously proposed TD based unsupervised FE~\cite{10.1007/s13538-011-0052-z} since how to select features is quite similar. 
Suppose that {\color{black}$j$ and $k$} are attributed to samples and {\color{black}$i$ is} attributed to features. For example, $j$ {\color{black}represents the} distinction {\color{black} between genders}  (e.g., $j\leq \frac{M}{2}$ is female and $j > \frac{M}{2}$ is male) and $k$ {\color{black}represents} age
(e.g.,  $k \leq \frac{K}{2}$ is young and $k > \frac{K}{2}$ is old) and $i$s are properties of persons (e.g., income, weight, height, blood pressure and so on). Then $x_{ijk}$ is the value of $i$th property of the person labeled by $j,k$.

After getting TD, we can find some $u_{\ell_2 j} $ which is distinct between genders and $u_{\ell_3 k}$ which is distinct between ages. 
Next, by investigating $|G(\ell_1 \ell_2 \ell_3)|$  we can find which $\ell_1$s have larger contribution. {\color{black}A}ssuming $u_{\ell_1 i}$ selected  by the  $\ell_2$ and $\ell_3$ obey Gaussian (Null hypothesis), we can attribute $P$-values to $i$s as
\begin{equation}
    P_i = P_{\chi^2} \left [ > \sum_{\ell_1} \left (\frac{u_{\ell_1 i}}{\sigma_{\ell_1}} \right)^2 \right] \label{eq:Pi}
\end{equation}
where the summation is taken over only the selected $\ell_1$s, $P_{\chi^2}[>x]$ is the cumulative $\chi^2$ distribution and $\sigma_{\ell_1}$ is the standard deviation optimized such that the distribution of the resulting $P$-values is coincident with the Gaussian distribution as much as follows as shown in the below. $P$-values are corrected by BH {\color{black} (Benjamini-Hochberg)} criterion and $i$s associated with adjusted $P$-values less th{\color{black}a}n the threshold value, say 0.05, are selected. 

In order to optimize SD, first we compute histogram of $1-P_i$ as $h_s(1-P)$ that is the frequency of $i$s in the $s$th bin excluding features to be selected, i.e., those associated with adjusted $P$-values less than the threshold value, say 0.01. 
Then we compute standard deviation of $h_s$ as
\begin{eqnarray}
\langle h_s \rangle &=& \frac{1}{S} \sum_{s=1}^S h_s \\
\sigma_{h} & =& \sqrt{\frac{1}{S} \sum_{s=1}^S \left ( h_s - \langle h_s \rangle \right)^2}
\end{eqnarray}
 where $S$ is the total number of bins.  Since $P$-values are computed {\color{black}as a} function of $\sigma_{\ell_1} $, there should be the optimal $\sigma_{\ell_1} $ that gives the minimum $\sigma_h$ supposed to be  most coincident with the null hypothesis that $u_{\ell_1 i}$ obeys Gaussian since $h_s(1-P_i)$ should be flat (i.e., $h_s(1-P_i) = \frac{N'}{S}$ where $N'$ is the total number of the not selected features) if the null hypothesis is totally true. 

When we analyze not a tensor but a matrix (i.e., sinusoidal data and RCS-GCM in the below), we can use singular value decomposition (SVD) 
\begin{equation}
x_{ij} = \sum_\ell \lambda_\ell u_{\ell i} u_{\ell j}
\end{equation}
instead of TD. Since $u_{\ell i}$ is always associated with $u_{\ell j}$ in SVD, we can decide which $u_{\ell i}$ should be used for the feature selection just after we decide which $u_{\ell j}$ is of interest.

\subsection{BTuD based unsupervised FE}

{\color{black}Up to} the stage {\color{black}at} which {\color{black}we decide which} $\ell_1$  to use to select $i$s, the processes are identical to those in TD based unsupervised FE. Then, $P$-values that  $u_{\ell_1 i}>(<)0$ when $ m_{u_{\ell_1 i}}<(>)0$ are attributed to $i$ as
\begin{equation}
    P_i = P_{\chi^2} \left [ > \sum_{\ell_1} \frac{ m_{u_{\ell_1 i}}^2}{ S_{\ell_1 \ell_1}}  \right] . \label{eq:Pi_BTD}
\end{equation}
The summation is taken over only the selected $\ell_1$s,
 $P$-values are corrected by BH criterion and $i$s associated with adjusted $P$-values less then the threshold value, say 0.05, are selected. 

 \subsection{Data sets}
\subsubsection{Synthetic data}

To test {\color{black}whether} the results of BTuD {\color{black}can be} used for feature selection,  we  prepared {\color{black} a tensor, $x_{ijk} \in \mathbb{R}^{N \times M \times K}$, as follows} 
\begin{equation}
x_{ijk} \sim \left \lbrace 
\begin{array}{lcc}
    \mathcal{N}(\mu,1) &  i \leq N_1 {\color{black} \leq N} , & j\leq \frac{M}{2}, k\leq \frac{K}{2} \\
    \mathcal{N} (0, 1) & \mbox{otherwise}
\end{array} \right.
\end{equation}
by which we can test {\color{black}whether} BTuD  based unsupervised FE can select $i \leq N_1$ correctly or not.

\subsubsection{Sinusoidal data}

To test if the results of BTuD is used for feature selection,  we also prepared {\color{black} a matrix, $x_{ij} \in \mathbb{R}^{N \times M}$, as} the following 
\begin{equation}
x_{ij} \sim \left \lbrace 
\begin{array}{lcc}
     \sin \left (  2 \pi \frac{j}{3} + \epsilon_i \right ),&  \epsilon_i \in   \mathcal{N} (0, 1), &  i \leq N_1 {\color{black} \leq N} \\
    \mathcal{N} (0, 1),  &   N_1 {\color{black} < i \leq N} .
\end{array} \right.
\end{equation}
by which we can test if BTuD  based unsupervised FE can select $i \leq N_1$ correctly or not.

\subsubsection{RCS-GCM}
GCM~\cite{PhysRevLett.65.1391} is a globally coupled system of multiple chaotic systems $x_{ij+1}=f(x_{ij},a)$, where $j$ is regarded {\color{black}as} time step,  formulated as follows.
\begin{eqnarray}
    x_{i j+1 }   &= &  (1-g) f(x_{ij},a) + \frac{g}{N} \sum_{i'=1}^N  f(x_{i'j}, a) \\
    f(x,a) &= & 1-ax^2
\end{eqnarray}
where $g$ is the coupling parameter and $a$ is the non-linearity parameter, the increase of which results in the chaotic behaviour of $x_{ij} {\color{black} \in \mathbb{R}^{N \times M}}$. 
By adjusting the parameters $a, g$ of this system, we can generate a globally synchronized state, a partially synchronized state with correlations between parts of dimensions, and a chaotic state in which all dimensions create an independent turbulent state. 
In this study, we assume for the data-generating model of this world that all dimensions have  different correlation lengths. 
The original GCM cannot achieve such a system. 
To generate  variables that are a mixture of long correlated ordered states and short correlated random states, we introduce GCM with random parameters as
\begin{eqnarray}
    x_{i j+1}   &= &  g_{ii} f(x_{ij},a_i) + \frac{1}{N} \sum_{i'=1}^N g_{ii'} f(x_{i'j}, a_{i'}), \\
    g_{ii'} & =& (1-c) \delta_{ii'} + c \epsilon_{ii'},   \\
    a_i & =& a +(1-a) \epsilon_i, \\
    f(x,a) &= & 1-ax^2,
\end{eqnarray}
where $\epsilon_{ii'}$ and $\epsilon_i$ are uniform random numbers as $\epsilon_i, \epsilon_{ii'} \sim [0,1]$.

This model extension has realized the behavior of the data generation model considered in this study, in which some dimensional groups behave synchronously with finite correlation lengths, whereas others behave chaotically with small correlation lengths. 
The specific parameters used in this study to generate data were  $a=1.75, c=0.04,$ and $N=10^4$, such that a single $f(x,a)$ falls in the chaotic region ($a > 1.48$). 
$c$, which expresses the strength of pairwise interactions between individual maps, is taken to be sufficiently small not to suppress the chaotic nature completely because of synchronization among individual maps and for $f(x,a)$ to have the mixture of ordered and random states. 
$j$s are taken to be $1 \leq j \leq 10^2$. Thus, the generated dataset is $x_{ijk} \in \mathbb{R}^{10^4 \times 10^2}$. 
Initial values ($x_{i0}$) are drawn from the same uniform distribution, $[0,1]$.

\subsubsection{Gene expression}

The gene expression profiles used in this study were downloaded
from the {\color{black}G}ene {\color{black}E}xpression {\color{black}O}mnibus (GEO) with GEO ID GSE142068.
Twenty{\color{black}-}four profiles named ``GSE142068\_count\_XXXXX.txt.gz'' were downloaded, where ``XXXXX'' indicates one of the 24 tissues, i.e., AdrenalG, Aorta, BM (Bone marrow), Brain, Colon, Eye, Heart, Ileum, Jejunum, Kidney, Liver, Lung, Pancreas, ParotidG, PituitaryG, SkMuscle, Skin, Skull, Spleen, Stomach, {\color{black}Testis}, Thymus, ThyroidG, and WAT (white adipose tissue), which were treated with 15 drugs: Alendronate, Acetaminophen, Aripiprazole, Asenapine, Cisplatin, Clozapine, Empagliflozin, Lenalidomide, Lurasidone, Olanzapine, Evolocumab, Risedronate, Sofosbuvir, and Teriparatide, and Wild type (WT).

They were formatted as tensor, $x_{ijkm} \in \mathbb{R}^{N \times 24 \times 18 \times 2}$, for $N$ genes, 24 tissues, 18 drug treatments, and two replicates. {\color{black}By} applying HOSVD {\color{black} (higher order singular value decomposition)} to $x_{ijkm}$, we get
\begin{equation}
x_{ijkm} = \sum_{\ell_1\ell_2\ell_3\ell_4}
G(\ell_1\ell_2\ell_3\ell_4) u_{\ell_1j}
u_{\ell_2k} u_{\ell_3m} u_{\ell_4 i} \label{eq:TD}
\end{equation}
where $G \in \mathbb{R}^{N \times 24 \times 18 \times 2}$ is the core tensor, 
$u_{\ell_1 j} \in \mathbb{R}^{24 \times 24}$, $u_{\ell_2 k} \in \mathbb{R}^{18 \times 18}$,$u_{\ell_3 m} \in \mathbb{R}^{2 \times 2}$, and $u_{\ell_4 i} \in \mathbb{R}^{N \times N}$, represents singular value matrices that are also orthogonal matrices. 
$x_{ijkm}$ is considered to be standardized as $\sum_i x_{ijkm} =0$ and $\sum_i x_{ijkm}^2 =N $.

\section{Results}\label{sec2}

\subsection{Performance test of BTuD based unsupervised FE}

Since the proposal to interpret Tucker decomposition as sets of linear regression is new, to {\color{black}determine whether} BTuD based unsupervised FE works correctly, we apply BTuD based unsupervised FE to the synthetic data sets with $N=1,000, M=K=20, N_1=10, \mu=1$. 
To get BTuD, we employed  results given by HOOI (Higher Order Orthogonal Iteration of tensors)~\cite{Taguchi2024} with {\color{black}the} setting ``max\_iter=500,tol=1e-8'' (we used \verb+tucker+ function in rTensor package~\cite{rTensor}). We also restricted  that $\ell_1 \leq 10$ and $\ell_2, \ell_3 \leq 5$.
HOOI successfully  converged to $m_{u_{\ell_1 i}}$, $m_{u_{\ell_2 j}}$,  and $m_{u_{\ell_3 k}}$ (Here we did not use the proposed Algorithm \ref{algorithm}).

One might wonder why we can judge that HOOI converged to the solution required for BTuD and decided to use HOOI instead of proposed (time{\color{black}-}consuming) algorithm, Algorithm \ref{algorithm}.
Since $m_{u_{\ell_1 i}}$, $m_{u_{\ell_2 j}}$,  and $m_{u_{\ell_3 k}}$ are the functions of $u_{\ell_1 i}, u_{\ell_2 j}, u_{\ell_3 k}$ {\color{black} as shown in} eq. (\ref{eq:m}), 
they must be decided in the self-consistent manner; this means that we can check if solutions computed by HOOI can satisfy the conditions  that the solution of BTuD must satisfy, based upon the self-consistency, $m_{u_{\ell_1 i}}=u_{\ell_1 i}$, $m_{u_{\ell_2 j}}=u_{\ell_2 j}$,  and $m_{u_{\ell_3 k}}=u_{\ell_3 k}$. As a result we found that HOOI can converge to the solution that satisfies self-consistency and decided that we replace HOOI with the proposed time consuming Algorithm \ref{algorithm}, also for the following examples other than the synthetic data. The results computed by HOOI are employed unless it is explicitly noticed that HOOI is not used.

At first, we consider $\ell_1=1$ and  compute $P_i$ with eq. (\ref{eq:Pi_BTD}).
The obtained 1,000 $P_i$ {\color{black}value}s are corrected by BH criterion and $i$s associated with adjusted $P$-values less than 0.05 are selected. As can be seen in Table \ref{table:conf}, BTuD based unsupervised FE almost completely selected $i \leq 10$ among 1,000 $i$s for one hundred ensembles. One might wonder why considering $\ell_1=1$ enables us to select $i \leq N_1$ in spite of the unsupervised method since we did not provide any information about the propert{\color{black}ies} of {\color{black}dataset} at all. 

We define
\begin{equation}
    Y_{\ell_1 jk} = \sum_{\ell_2=1}^{L_2} \sum_{\ell_3=1}^{L_3}
    G( \ell_1 \ell_2 \ell_3) u_{\ell_2 j} u_{\ell_3 k}.
\end{equation}
Thus, if  $u_{\ell_2 j}$ and $ u_{\ell_3 k}$ are coincident with $x_{ijk}, i \leq N_1$,  $Y_{\ell_1 jk}$ with larger absolute $G(\ell_1 \ell_2 \ell_3)$ is also coincident with $x_{ijk}, i \leq N_1$.
Since 
\begin{equation}
    x_{ijk} \simeq \sum_{\ell_1=1}^{L_1} Y_{\ell_1 jk}  u_{\ell_1 i},
\end{equation}
  $u_{\ell_1 i}$ should take larger absolute values only for
$i \leq N_1 $ as well such that $Y_{\ell_1 jk}$ is coincident with  $x_{ijk}, i \leq N_1$. This results in the smaller $P$-values  attributed to $i \leq N_1$ by eq. (\ref{eq:Pi_BTD}) and $i \leq N_1$ will be selected.

\begin{figure}[htb]
\centering
\includegraphics[width=0.3\linewidth]{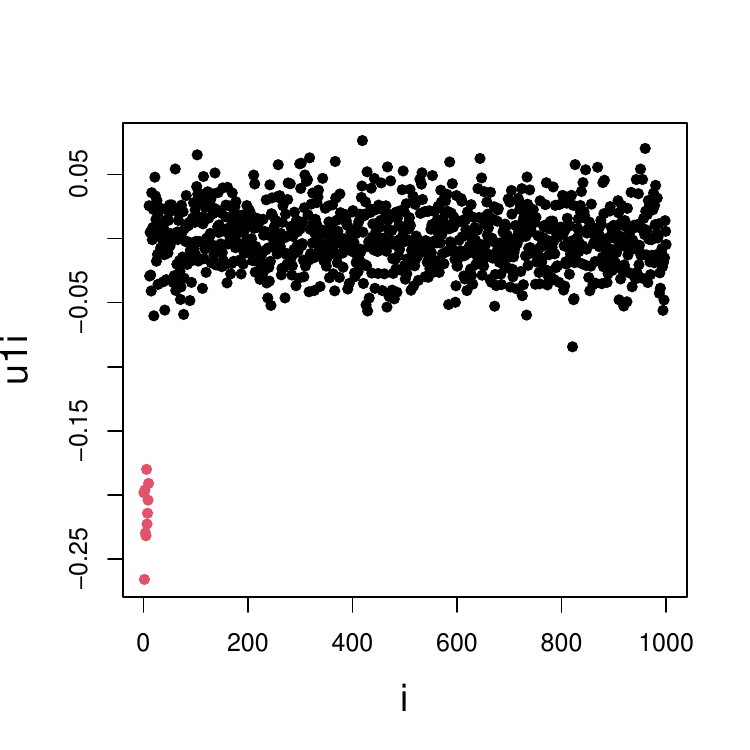}\includegraphics[width=0.6\linewidth]{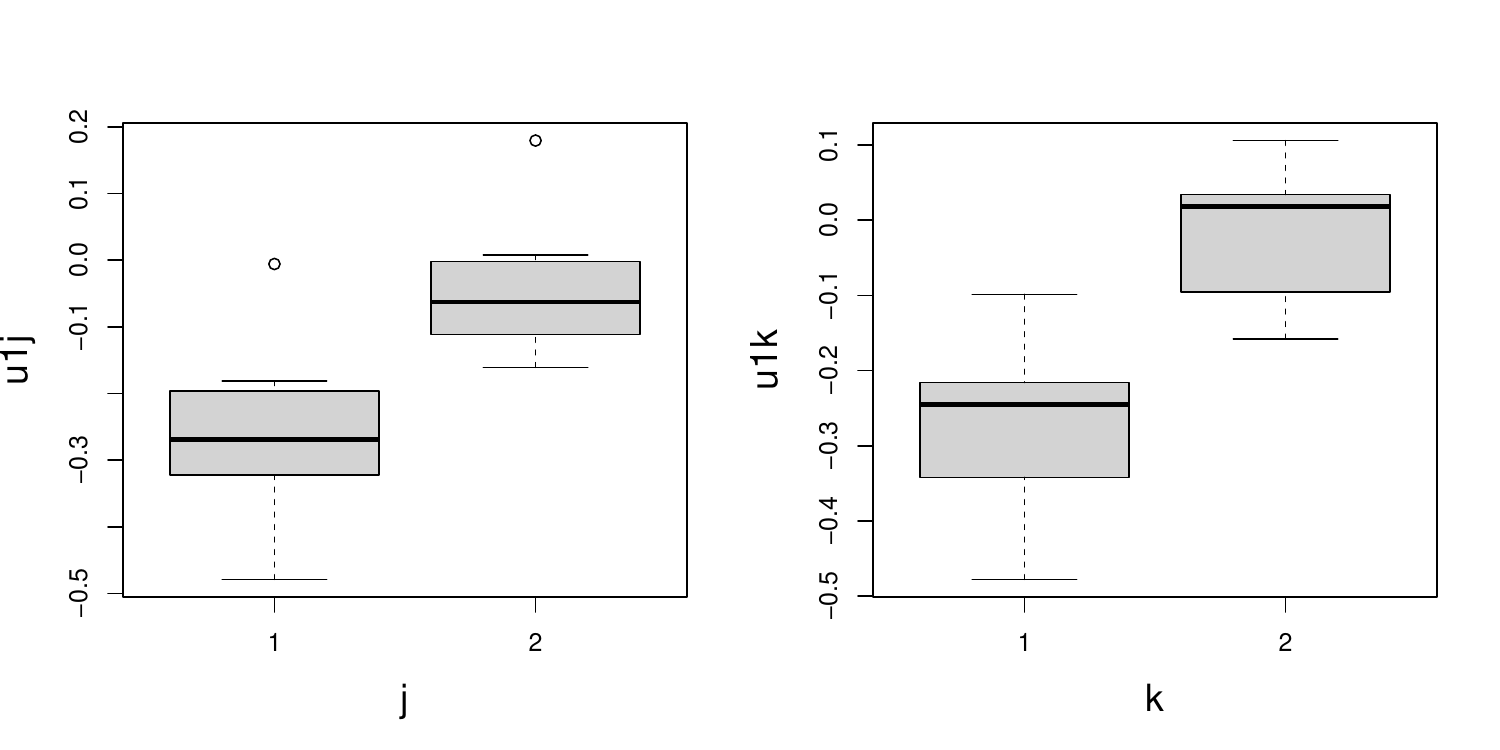}
\caption{ The results of BuTD applied to synthetic data. Left: $u_{1i}$. Red ones are $i \leq N_1$.
 middle: boxplot of $u_{1j}$ with 1 for $j\leq \frac{M}{2}$ and 2 for $j>\frac{M}{2}$, right:  boxplot of $u_{1k}$ with 1 for $k\leq \frac{K}{2}$ and 2 for $k>\frac{K}{2}$.   As
expected, since {\color{black} $x_{ijk}$, } $i \leq N_1$ are more likely represented as a product of two features, {\color{black}  $u_{1j}
u_{1k}$,} than others {\color{black} (i.e., $i > N_1$)} , {\color{black} $x_{ijk}$, $i \leq N_1$s } are associated  with larger absolute $u_{1i}$ {\color{black} because of larger absolute $G(111)$} and are accordingly associated with smaller $P$-values and selected.  }\label{fig:U121}
\end{figure}

In {\color{black}fact}, $Y_{1 jk}$ as well as  $u_{1 j}$ and $u_{1 k}$ are coincident with $x_{ijk}, i \leq N_1$.
For example, we checked if $u_{1 j}$ and $u_{1 k}$ are distinct between $j \leq \frac{M}{2}$, $k \leq \frac{K}{2}$ and $j > \frac{M}{2}$, $k > \frac{K}{2}$, which means that  $u_{1 j}$ and $u_{1 k}$ are coincident with $x_{ijk}, i \leq N_1$.  As can be seen in Table \ref{table:coin},
in {\color{black}at least} 85 out of 100 ensembles, they are significantly distinct and $G(1 1 1)$ takes larger absolute values (not shown here), which means that $Y_{1 jk}$ is supposed to be coincident with $x_{ijk}, i \leq N_1$ as well.
In addition to this, we also checked if $u_{1 j} u_{1 k}$s themselves are distinct between $j \leq \frac{M}{2}, k
\leq \frac{K}{2}$ and others. Then as can be seen in Table \ref{table:coin2}, more than 90 out of 100 ensembles, they are distinct; in the remaining 10 out of 100 ensembles, we can also find $Y_{1 jk}$s are distinct  between $j \leq \frac{M}{2}, k
\leq \frac{K}{2}$ and others with considering $\ell_2 >1$ and $\ell_3 >1$ since not only $G(1 1 1)$ but also other $G(1 \ell_2 \ell_3)$s have negligible contributions as well. Thus the reason why absolute values of $u_{1i}$s for $i \leq N_1$ are much larger than those for $i > N_1$ is because $u_{1 j}u_{1 k}$s or
$Y_{1 jk}$s are distinct between $j \leq \frac{M}{2}, k
\leq \frac{K}{2}$ and others (Fig. \ref{fig:U121}).

Why are $(u_{1 j}, u_{1 k})$s or
$Y_{1 jk}$s distinct between $j \leq \frac{M}{2}, k
\leq \frac{K}{2}$ and others {\color{black}despite the} unsupervised nature? 
When the number of $u_{\ell_1 i}$, $u_{\ell_2 j}$ and $u_{\ell_3 k}$ are restricted, 
it is better to select those coincident with the majority of $x_{ijk}$ to minimize residuals.  Since $x_{ijk} , i \leq N_1$ share the distinction between $j \leq \frac{M}{2}, k \leq \frac{K}{2}$ and others whereas $x_{ijk}, i >N_1$ have nothing to share since they are composed of pure random numbers, it is better to select those coincident with $x_{ijk} , i \leq N_1$ and $u_{\ell i} , i \leq N_1$  have larger absolute values, which results in smaller $P$-values attributed to $u_{\ell_1 i} , i \leq N_1$  that are finally selected.

 \begin{table}[h]
\caption{Confusion matrix of  BTuD based unsupervised FE applied to synthetic data. $i \leq N_1$ are associated with distinction between two classes and are supposed to be selected. Those associated with adjusted $P$-values less than 0.05 are selected.  $i \leq N_1$ and  the selected ones are consistent well.} \label{table:conf}%
\begin{tabular}{@{}llll@{}}
\toprule
  & \multicolumn{3}{c}{BTuD based unsupervised FE} \\
 &   & $i>N_1$ & $i \leq N_1$\\
\midrule
Adjusted $P$-value  & $>0.05$   & 990 & 0.09  \\
   &  $\leq 0.05$  & 0  & 9.91 \\
\botrule
\end{tabular}
\end{table}

\begin{table}[h]
\caption{ The confirmation if  $u_{1j}$ as well as $u_{1k}$ are consistent with the distinction between two classes. The number of ensembles where $u_{1j}$ or $u_{1k}$ is consistent with  the distinction between $j \leq \frac{M}{2}$ and $j > \frac{M}{2} $ or $ k
\leq \frac{K}{2}$  and $ k
 >\frac{K}{2}$, respectively.  In the majority of ensembles, $u_{1j}$ as well as $u_{1k}$ are consistent with the distinction between two classes. $P$-values are computed by \textit{t} test. }\label{table:coin}%
\begin{tabular}{@{}lllllll@{}}
\toprule
   & \multicolumn{2}{c}{$u_{1j}$ }  & \multicolumn{2}{c}{$u_{1k}$ }\\
  Adjusted $P$-value  & $>0.05$   &  $\leq 0.05$ & $>0.05$   &  $\leq 0.05$\\
\midrule
  The number of ensembles & 15 & 85 & 12 & 88  \\
\botrule
\end{tabular}
\end{table}

\begin{table}[h]
\caption{ The confirmation if $(u_{1j}, u_{1k})$ is consistent with the distinction between two classes.  The number of ensembles where $(u_{1j}, u_{1k})$ is coincident with  the distinction between $j \leq \frac{M}{2}$ , $ k \leq \frac{K}{2}$  and others.  In the majority of ensembles, $(u_{1j}, u_{1k})$ is consistent with the distinction between two classes. $P$-values are computed by \textit{t} test. }\label{table:coin2}%
\begin{tabular}{@{}lllllll@{}}
\toprule
   & \multicolumn{2}{c}{$(u_{1j}, u_{1k})$ }\\
  Adjusted $P$-value  & $>0.05$   &  $\leq 0.05$ \\
\midrule
  The number of ensembles & 8& 92   \\
\botrule
\end{tabular}
\end{table}

Although one might wonder whether BTuD based unsupervised FE might be useless unless there are majority features as in the above synthetic data, it is not the case. To show that BTuD based unsupervised FE is effective even if there are no majority features, we applied BTuD based unsupervised FE to $x_{ijk}$ that includes sinusoidal function{\color{black}s} with distinct phases as a part
($N=10,000, M=100,K=1, N_1=1,000$). We also restricted  that $\ell_1 \leq 10$ and $\ell_2 \leq 2, \ell_3 \leq 1$. Since sinusoidal functions with distinct phases are distinct with each other and cannot be composed of majority features, one might think BTuD based unsupervised FE cannot detect them. In spite of the concern, BTuD based unsupervised FE successfully select{\color{black}s} a set of sinusoidal functions ($i \leq N_1$) among 10,000 features (Table \ref{table:sin}) when we consider $\ell_1=1,2$ and  compute $P_i$ with eq. (\ref{eq:Pi_BTD});
the obtained 10,000 $P_i$s are corrected by BH criterion and $i$s associated with adjusted $P$-values less than 0.05 are selected.

\begin{figure}[htb]
\centering
\includegraphics[width=0.3\linewidth]{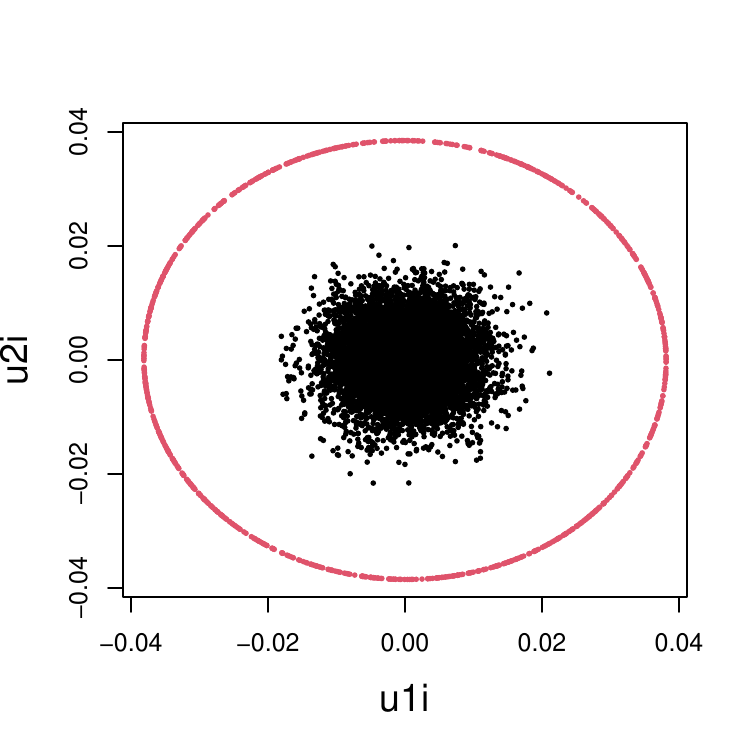}\includegraphics[width=0.6\linewidth]{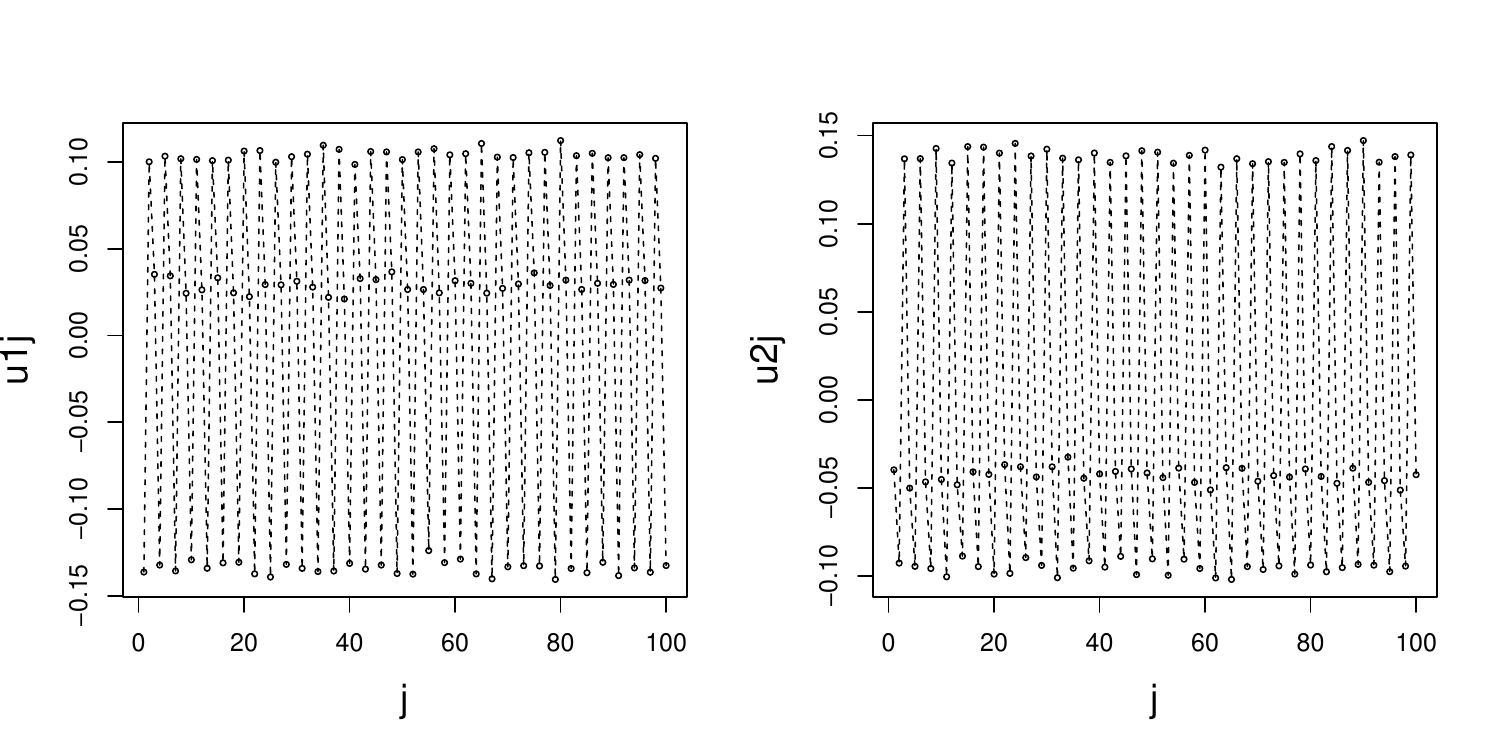}
\caption{The results of BuTD applied to sinusoidal data. Left: $u_{1i}$  vs $u_{2i}$. Red ones are $i \leq N_1$. As
expected, since $x_{ij}, i \leq N_1$s that are sinusoidal are more likely represented as a linear combination of two sinusoidal features, $u_{1j}$  (middle) and $u_{2j}$ (left),
$u_{1i} \propto \sum_j u_{1j} x_{ij}$ and $u_{2i} \propto \sum_j u_{2j} x_{ij}$ should have larger absolute values than others ($i > N_1$). Then $u_{1i}$ and $u_{2i}$ with $i \leq N_1$ are associated  with   smaller $P$-values and selected.  }\label{fig:U12}
\end{figure}

The reason why it could occur is as follows; since sinusoidal functions, $A \sin (x + \delta)$,  with distinct phases, $\delta$, that correspond to $x_{ij}, i\leq N_1$ can be represented as a linear combination of $\sin$ function and $\cos$ function, 
\begin{equation}
    A \sin (x + \delta) = A \cos \delta \sin x + A \sin \delta \cos x,
\end{equation}
majority features that are sinusoidal and allow us to select  $x_{ij}, i\leq N_1$  are represented as well with $u_{1j}$ and $u_{2j}$ if $u_{1j}$ and $u_{2j}$  correspond to $\sin$ and $\cos$, respectively. As can be seen  in Fig. \ref{fig:U12}, $u_{1j}$ and $u_{2 j}$ are sinusoidal. Since we use SVD instead of TD and SVD makes $u_{\ell j}$ to directly correspond to $u_{\ell i}$,  we can select $x_{ij}, i \leq N_1$ by using $u_{1i}$  and $u_{2i}$ to select $i$s.

\begin{table}[h]
\caption{Confusion matrix of BTuD applied to sinusoidal data averaged over 100 ensembles. $i \leq N_1$ are associated with sinusoidal function and are supposed to be selected. Those associated with adjusted $P$-values less than 0.05 are selected.  $i \leq N_1$ and  the selected ones are consistent well.} \label{table:sin}%
\begin{tabular}{@{}llll@{}}
\toprule
  & \multicolumn{3}{c}{BTuD based unsupervised FE} \\
 &   & $i>N_1$ & $i \leq N_1$\\
\midrule
Adjusted $P$-value  & $>0.05$   & 8999.75 & 0.00 \\
   &  $\leq 0.05$  & 0.25  & 1000 \\
\botrule
\end{tabular}
\end{table}

\begin{figure}[!htb]
\centering
\includegraphics[width=0.4\linewidth]{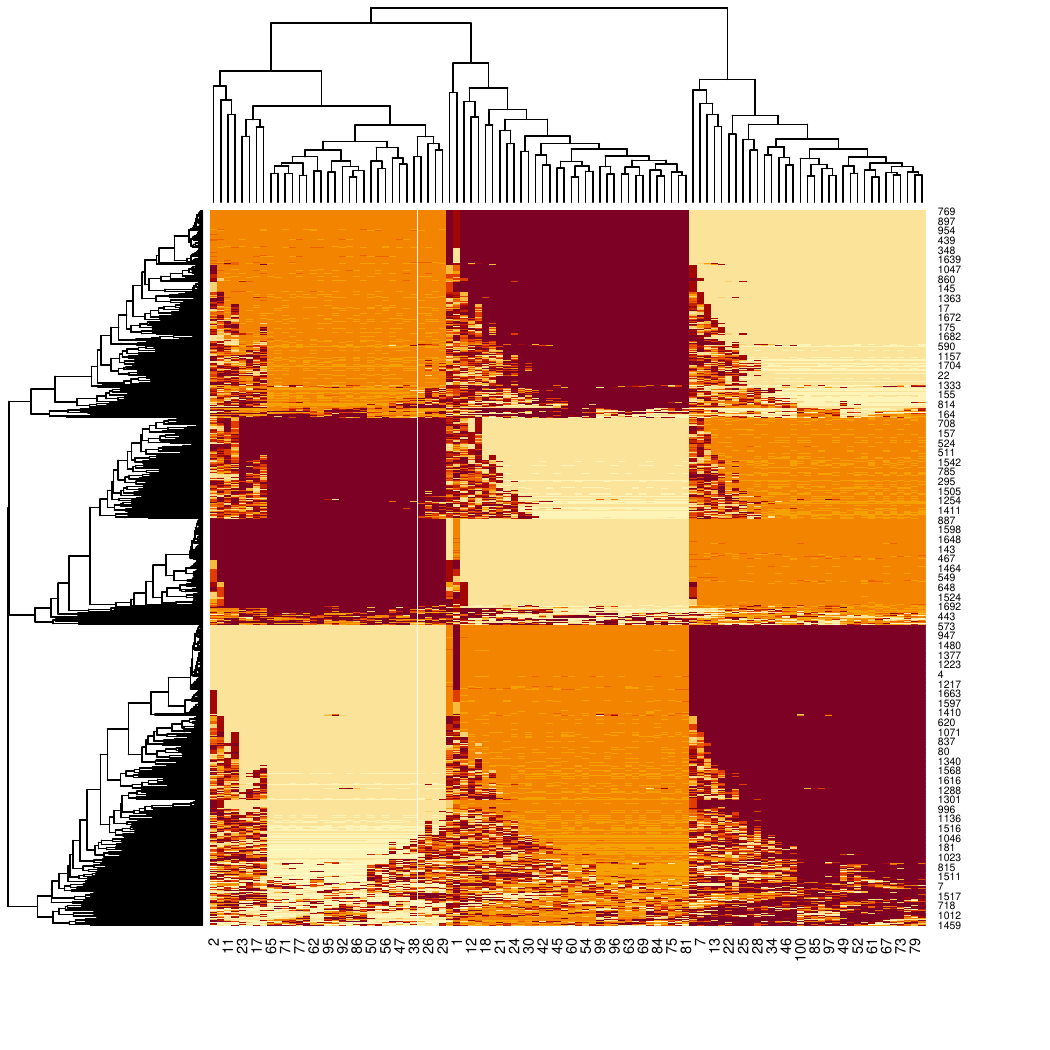}\includegraphics[width=0.4\linewidth]{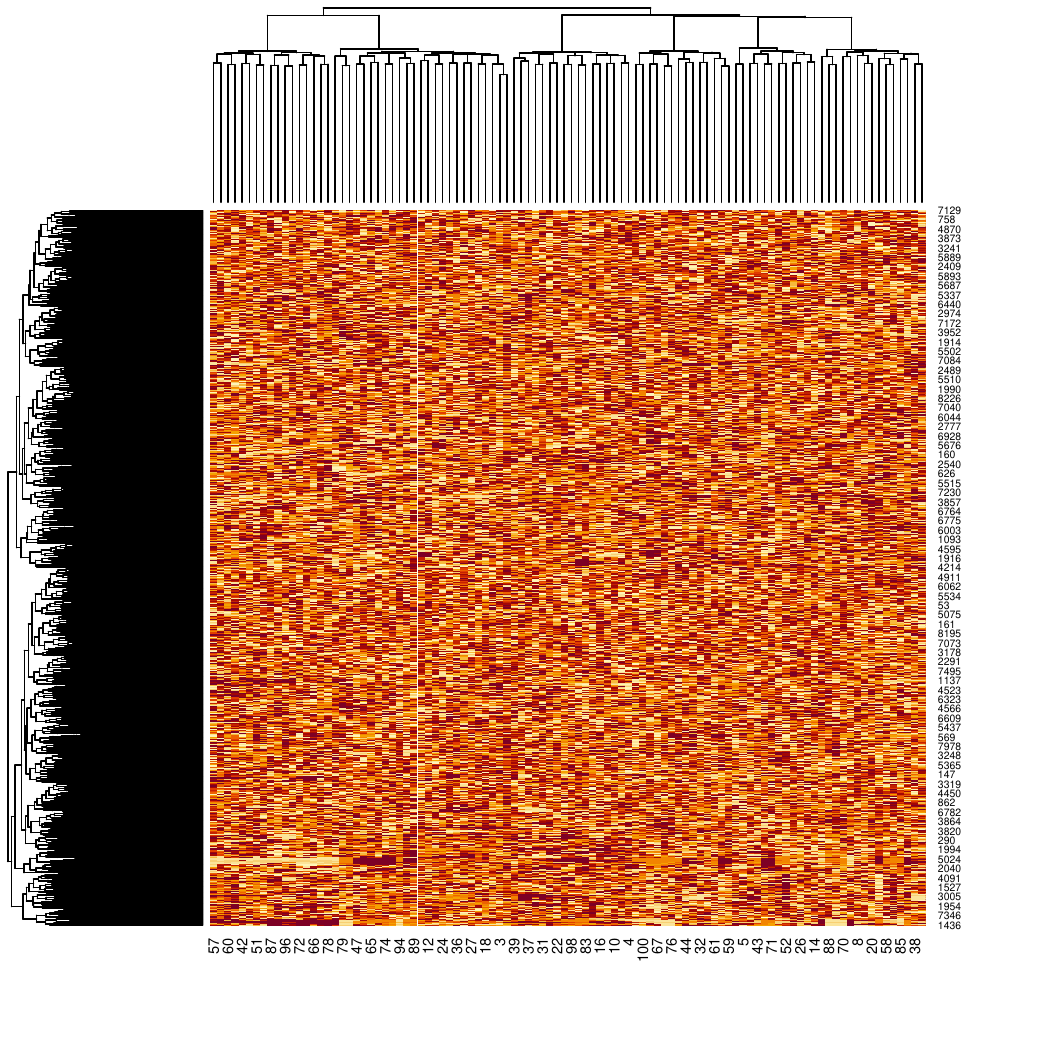}
\caption{Heatmap of $x_{ij}$ for 1707 $i$s selected (left)  and 8293 $i$s not selected (right) by BTuD based unsupervised FE, respectively (Table \ref{table:RCS-GCM}). 
Rows ($i$) and columns ($j$) are clustered by the hierarchical clustering. 
It is obvious that $i$s with three states are selected  and not selected $i$s lack ordered state. 
}\label{fig:RCS-GCM}
\end{figure}

\begin{figure}[!htb]
\centering
\includegraphics[width=0.5\linewidth]{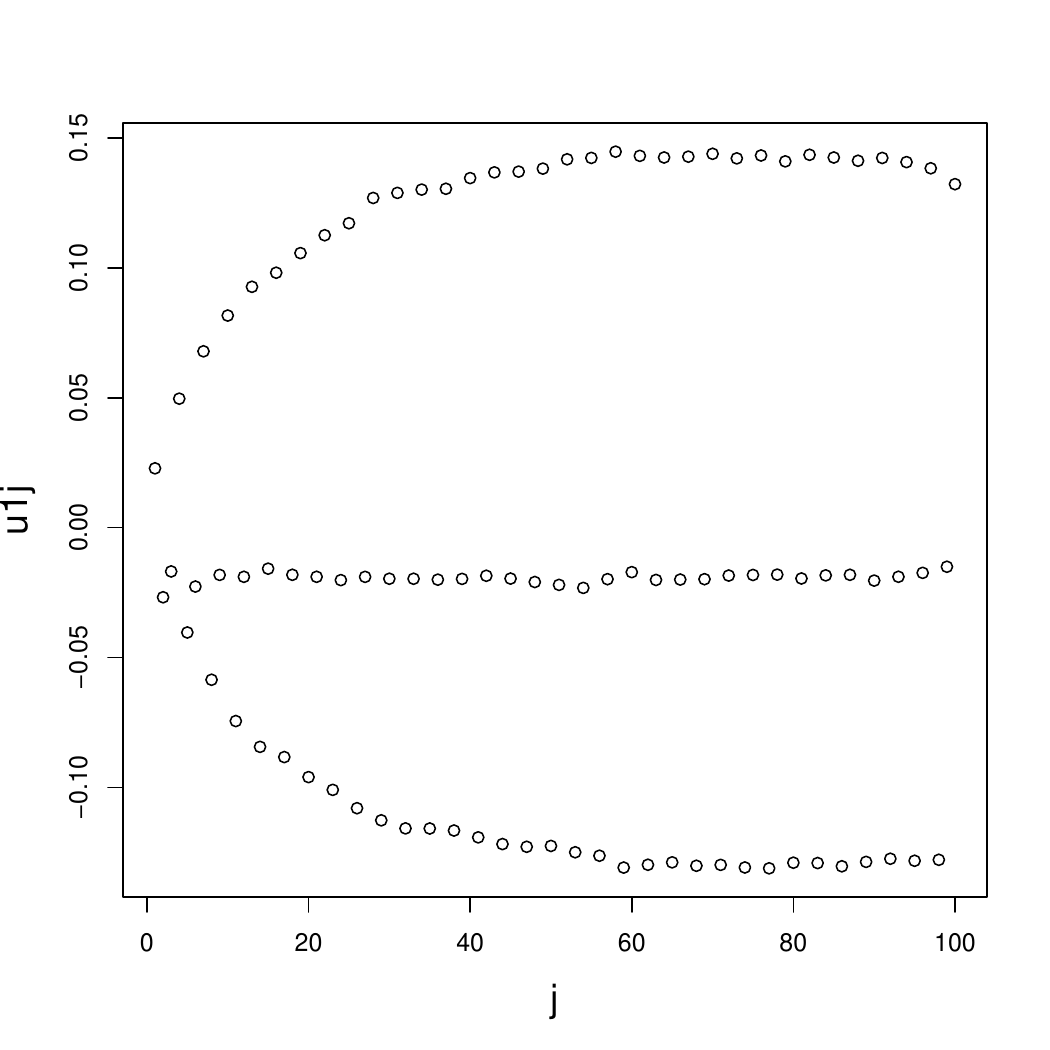}
\caption{ The result of BTD applied to RCS-GCM. Vertical axis is $u_{1j}$ and horizontal axes is $j$.} \label{fig:SNU}
\end{figure}

In the above two examples, we have shown that BTuD based unsupervised FE can be successful when we know what the correct answer is. In the following, we applied BTuD based unsupervised FE to two cases, {\color{black}whose} correct answers  we do not know in advance; one is the result of numerical model and another is that of real data, gene expression profiles.
The first (numerical) example is so{\color{black}-}called Randomized Coupling Strength GCM (RCS-GCM)~\cite{mototake2023signalidentificationsignalformulation}. {\color{black}By} tuning  parameters,  RCS-GCM can generate the mixture of ordered (periodic) states and random (chaotic) states. The performance test of BTuD based unsupervised FE applied to RCS-GCM is whether BTuD based unsupervised FE can,  in {\color{black} a} fully unsupervised manner, distinguish ordered states from random states in RCS-GCM.  As can be seen in Table \ref{table:RCS-GCM}, using $P_i$ computed with eq. (\ref{eq:Pi_BTD}) with setting $\ell_1=1$, we can identify many $i$ {\color{black}values} which are apparently composed of three states (Fig. \ref{fig:RCS-GCM}) in spite of that we  intentionally select neither ordered state nor three states at all.    
The reason why three{\color{black}-}state variables can be selected in the unsupervised manner is because $u_{1j}$ represents the three state vector that enables us to select $x_{ij}$s associated with three state unintentionally (Fig. \ref{fig:SNU}). Since here we again analysed not a tensor but a matrix, correspondence between $u_{\ell_1 i}$ and $u_{\ell_2 j}$ is again always that $\ell_1=\ell_2$, since $G(\ell_1 \ell_2)$ is always diagonal.
Here we would like to emphasize that the separation between ordered states and others in the present study is much better than the previous one~\cite{mototake2023signalidentificationsignalformulation} where TD based unsupervised FE was employed.
Thus, we can conclude that BTuD based unsupervised FE can deal with the data sets,  correct answers of which we do not know. 

\begin{table}[!htb]
\caption{The performance of BTuD based unsupervised FE applied to RCS-GCM.  
The 1707 $i$s are associated with adjusted $P$-values less than  0.05 and are s{\color{black}e}lecteed. c{\color{black}o}nsistent with ordered state (see Fig. \ref{fig:RCS-GCM})}\label{table:RCS-GCM}%
\begin{tabular}{@{}lllllll@{}}
\toprule
   & \multicolumn{2}{c}{$(u_{1j}, u_{1k})$ }\\
  Adjusted $P$-value  & $>0.05$   &  $\leq 0.05$ \\
\midrule
  The number of $i$s & 8293 & 1707    \\
\botrule
\end{tabular}
\end{table}

\begin{figure}[!htb]
\centering
\includegraphics[width=\linewidth]{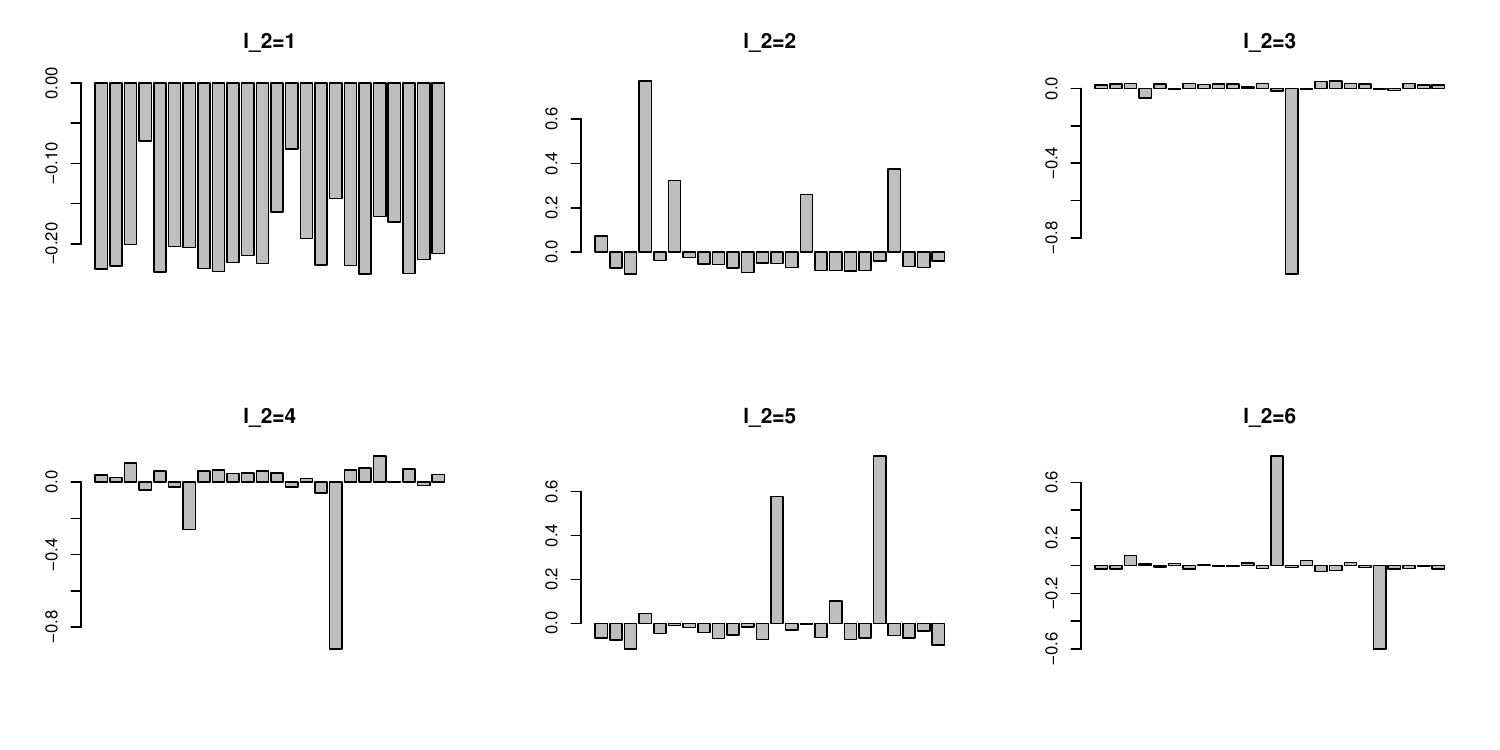}
\caption{ The results of BuTD applied to gene expression: $u_{\ell_2j}$. Horizontal axes  are 24 tissues. 
Top left: $\ell_2=1$, top middle: $\ell_2=2$, top right: $\ell_2=3$, bottom left $\ell_2=4$, bottom middle: $\ell_2=5$, bottom right: $\ell_2=6$. 
Please notice that $\ell_2=5,6$ are associated with the distinct combination of two common tissues. }\label{fig:tissue}
\end{figure}

\begin{figure}[!htb]
\centering
\includegraphics[width=\linewidth]{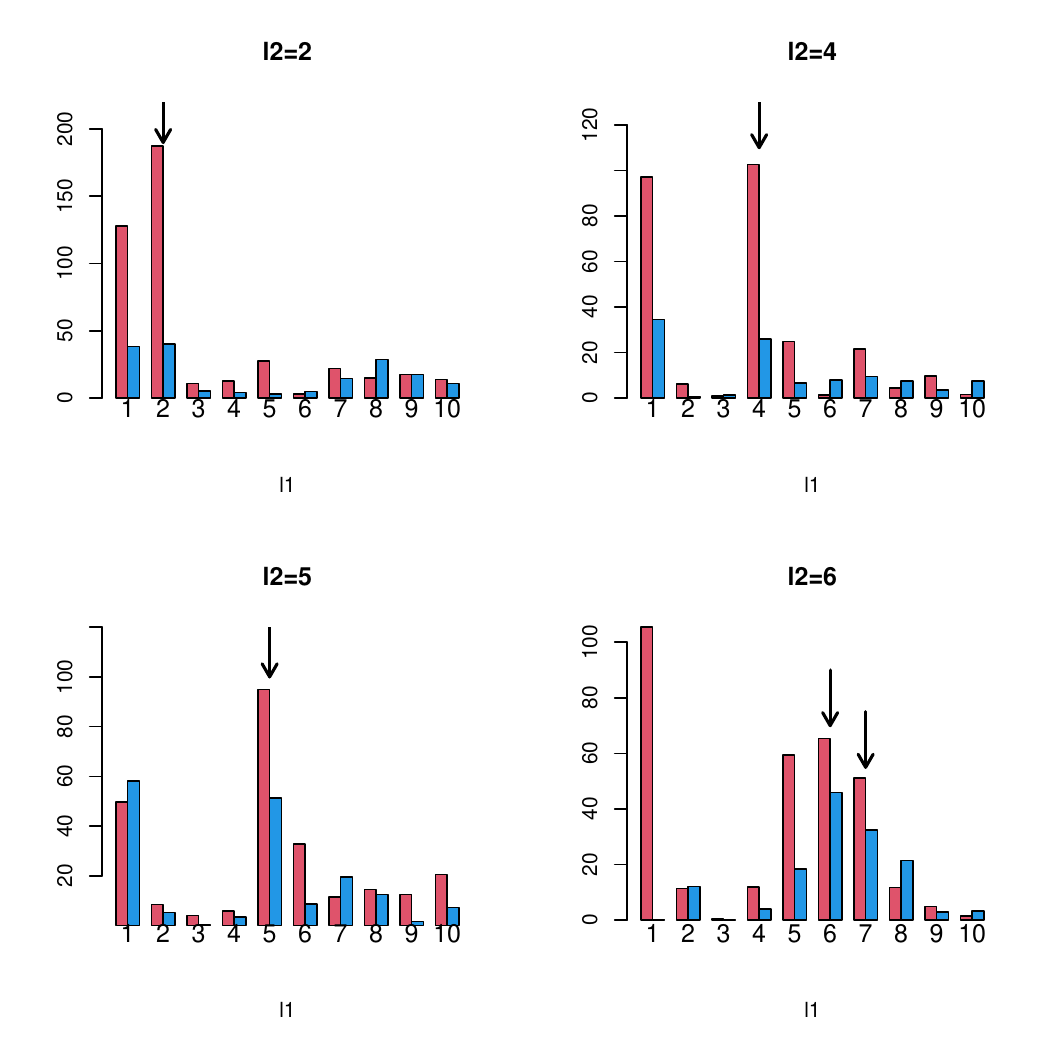}
\caption{The results of BuTD applied to gene expression:$|G(\ell_1,\ell_2,\ell_3,1)|$. Horizontal axes  are $\ell_1$. 
Top left: $\ell_2=2$,  top right: $\ell_2=4$, bottom left $\ell_2=5$,  bottom right: $\ell_2=6$.
Vertical arrows show the selected $\ell_1$s. Red: $\ell_3=2$, blue: $\ell_3=3$.} \label{fig:G}
\end{figure}

The second real example is gene expression profiles~\cite{KOZAWA2020100791}. With applying BTuD based unsupervised FE to gene expression profiles, we can identify not one but multiple sets of genes each of which is associated with distinct properties (tissue specificit{\color{black}ie}s) as identified previously~\cite{10.3389/fgene.2020.00695,10.1038/s41598-022-21474-z}.  
Although the following explanation is almost the replication of the previous study~\cite{10.3389/fgene.2020.00695}, since the employed method is not TD based unsupervised FE but BTuD based unsupervised FE in the present study, we briefly explain the procedure.
At first, we need to identify which $u_{\ell_2 j}$s are associated with tissue specificity.
Fig. \ref{fig:tissue} shows the $u_{\ell_2 j},  1\leq \ell_2 \leq 6$. 
Although $u_{1j}$ does not exhibit clear tissue-specificity, other five $u_{\ell_2 j}$s have clear tissue specific expression. Among them, we selected
$\ell_2=2,4,5,6$ that simultaneously exhibit specificities for multiple tissues (Please notice that $\ell_2=5,6$ are associated with the combination of two common tissues).
Next we found that $u_{2 k}$ and $u_{3 k}$ exhibit distinction between drug treatments and controls and $u_{1 m}$ exhibits independence of two replicates (not shown here).
Thus we can seek $|G(\ell_1,\ell_2,\ell_3,1)|, \ell_2=2,4,5,6, \ell_3=2,3$ to find which $\ell_1$s are associated with tissue specificities and distinction between drug treatments and 
controls, simultaneously (Fig. \ref{fig:G}). The selected $\ell_1$s are used to attribute $P$-values to $i$ using eq. (\ref{eq:Pi_BTD}). 

\begin{figure}[!htb]
\centering
\includegraphics[width=\linewidth]{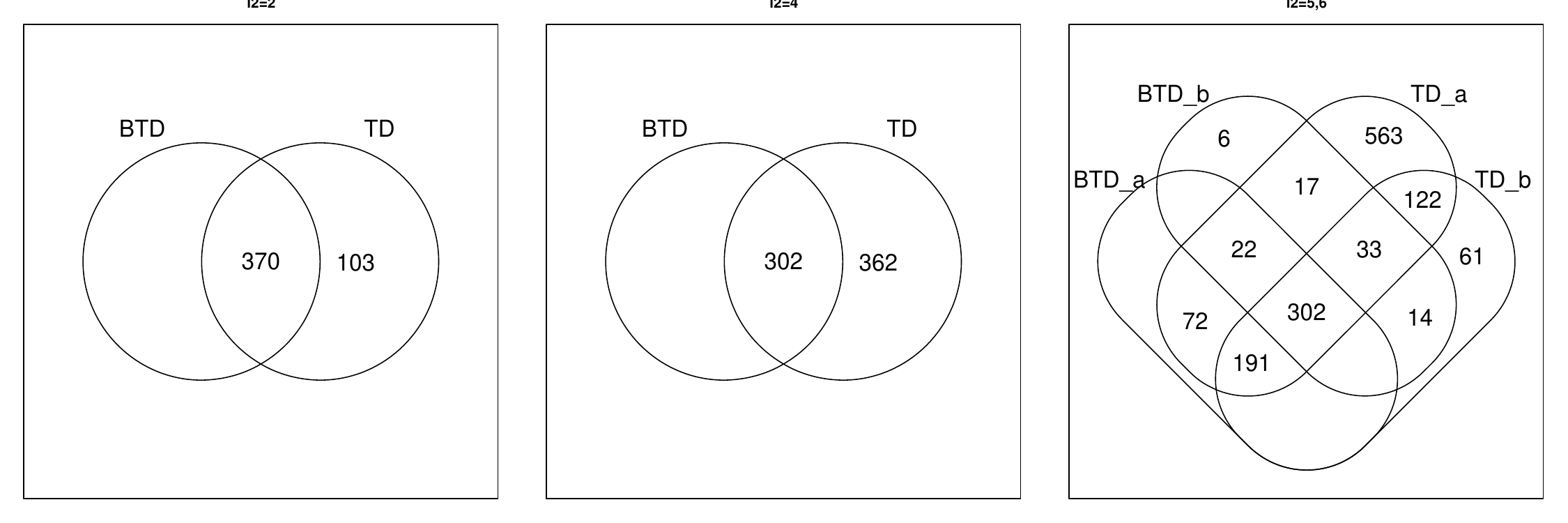}
\caption{Venn diagram of the selected genes between BTuD based unsupervised FE (present study) and TD based unsupervised FE (previous study~\cite{10.1038/s41598-022-21474-z}).
Left: $\ell_2=2$, middle: $\ell_2=4$, right:$\ell_2=5,6$. $\ell_2=5$  and  $\ell_2=6$ correspond to BTuD/TD\_a and BTuD/TD\_b, respectively.  Since $\ell_2=5,6$ are associated with the combination of two common tissues, they are treated together.}\label{fig:venn}
\end{figure}

Table \ref{table:gene} shows the number of selected genes.
\begin{table}[!htb]
\caption{The number of selected genes associated with tissue specificities and distinction between drug treatments and controls }\label{table:gene}%
\begin{tabular}{@{}lllllll@{}}
\toprule
   Adjusted $P$-value  & $>0.05$   &  $\leq 0.05$ \\
\midrule
 $\ell_2=2$   & 24052 & 369    \\
 $\ell_2=4$   & 24120 & 301 \\
 $\ell_2=5$   & 23834  & 587   \\
 $\ell_2=6$   & 24028  &  393     \\
\botrule
\end{tabular}
\end{table}
To validate genes selected by BTuD based unsupervised FE, we compare the  genes selected by BTuD based unsupervised FE with those selected by TD based unsupervised FE~\cite{10.1038/s41598-022-21474-z}. Since genes selected by TD based unsupervised FE were biologically evaluated and proven to be reasonable, if the overlap between genes selected by BTuD based unsupervised FE and those selected by TD based unsupervised FE is significant,  we can judge that BTuD based unsupervised FE selected reasonable genes.
Figure \ref{fig:venn} shows the Venn diagram between genes selected by BTuD based unsupervised FE and those selected by TD based unsupervised FE; since the former is almost always the subset of the latter, the overlap between them is definitely significant (Since $\ell_2=5,6$ are associated with the combination of two common tissues, they are grouped together).
Thus, BTuD based unsupervised FE is not only applicable to real data sets, but also can deal with data sets including sets of variables associated with multiple properties.

In conclusion, BTuD based unsupervised FE can be {\color{black}applied} to {\color{black}a} wide range of data sets without pre-assigned labels or {\color{black} prior} knowledges.


\section{Discussion}
\subsection{Equivalence between BTuD based unsupervised FE and TD based unsupervised FE}

Since BTuD-based unsupervised FE is practically {\color{black}identical to} TD-based unsupervised FE especially in the last {\color{black}example}, gene expression, it is important to clarify the reason of this coincidence; TD-based unsupervised FE is known to be successfully applied to many problems~\cite{Taguchi2024}.

\pgfmathdeclarefunction{gauss}{3}{%
  \pgfmathparse{0.7/(#2*sqrt(2*pi))*exp(-((x-#1)^2)/(2*#2^2))+#3}%
}
\begin{figure}[!htb]
\pgfplotsset{width=5cm}
\pgfplotsset{height=10cm}
\begin{tikzpicture}
\begin{axis}[every axis plot post/.append style={
  mark=none,domain=-1:2,samples=50}, 
  ylabel=$i$,xlabel =$u_{\ell_1 i}$,
  ymin=0,title=BTuD based unsupervised FE,
  axis x line*=bottom, 
  axis y line*=middle, 
  enlargelimits=upper] 
   \addplot[blue,mark=o,only marks] coordinates{(0.1,3)(0.2,4)(0.05,5)(0.05,6)};
  \addplot[red,mark=o,only marks] coordinates{(0.76,1)(0.70,2)};
  \addplot[blue] {gauss(0.05,0.375,6)};
  \addplot[blue] {gauss(0.05,0.375,5)};
  \addplot[blue] {gauss(0.2,0.375,4)};
  \addplot[blue] {gauss(0.1,0.375,3)};
   \addplot[red] {gauss(0.76,0.375,2)};
  \addplot[red] {gauss(0.70,0.375,1)};
\end{axis}
\end{tikzpicture}
\begin{tikzpicture}
\begin{axis}[every axis plot post/.append style={
  mark=none,domain=-1:2,samples=50}, 
  ylabel=$i$,xlabel =$u_{\ell_1 i}$,
  ymin=0,title=TD based unsupervised FE,
  axis x line*=bottom, 
  axis y line*=middle, 
  enlargelimits=upper] 
  \addplot[white,mark=o,only marks] coordinates{(0.05,7)};
  \addplot[blue,mark=o,only marks] coordinates{(0.1,3)(0.2,4)(0.05,5)(0.05,6)};
  \addplot[red,mark=o,only marks] coordinates{(0.76,1)(0.70,2)};
  \addplot[blue] {gauss(0,0.375,0)};
   \addplot[red] {gauss(0.75,0.375,0)/2};
\end{axis}
\end{tikzpicture}
\caption{The schematic figure that explains the equivalence between BTuD based unsupervised FE (left) and TD based unsupervised FE (right). Blue and red dots corresponds to $u_{\ell_1 i}$ attributed to not selected and selected features, respectively. For BTuD based unsupervised FE, Gaussian distributions having the common SD and distinct means are attributed to features to be selected (red) and those not to be selected (blue), respectively. Since the latter has zero within the 95 \% confidence intervals, they are rejected whereas the former is not rejected since zero is outside the 95 \% confidence interval. For TD based unsupervised FE, no Gaussian distribution is attributed to individual features, but two Gaussian distribution that have distinct means between features not to be selected (blue) and those to be selected (red) are assumed to exist.  The latter is decided if they are regarded to be outliers based upon Gaussian distribution attributed to features not to be selected as a whole (blue). Since SD is common for all the Gaussian distributions, the same $P$-values are attributed to  individual features in both methods and the same features are selected in both method.}\label{fig:exp}
\end{figure}
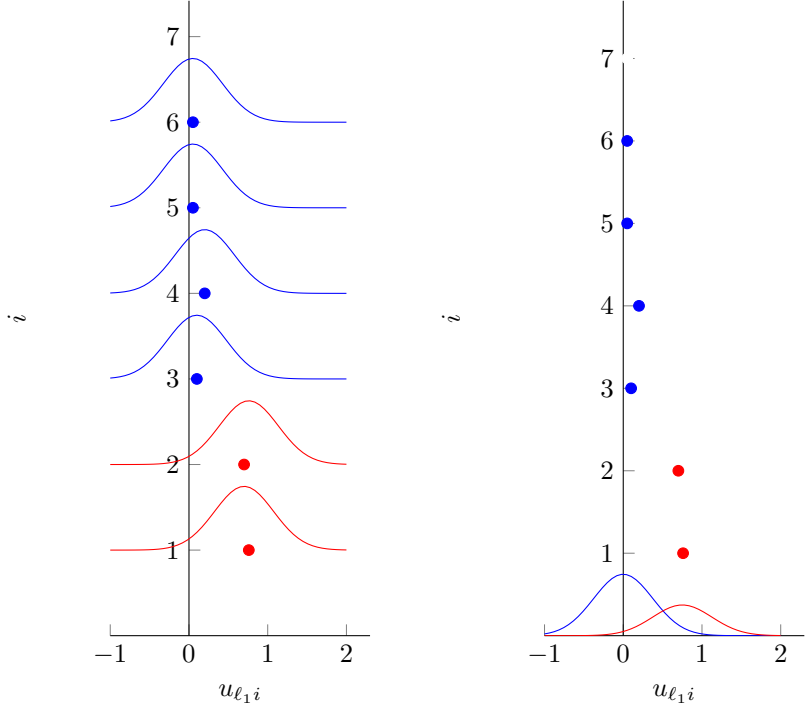

$P$-values are attributed to features with eqs. (\ref{eq:Pi}) and (\ref{eq:Pi_BTD})  for TD based unsupervised FE and BTuD based unsupervised FE, respectively. In TD based unsupervised FE, mean and SD are estimated based upon the distribution of  not selected features that are assumed to obey  Gaussian distribution in BTuD based unsupervised FE. Thus Gaussian distribution attributed to not selected features are empirically identical and thus the same $P$-values are attributed to not selected features in both methods.
The Gaussian distributions with common SD are attributed  to the selected feature{\color{black}s} in both method{\color{black}s}, but they are centered distinctly.
However, if we attribute $P$-values of zero to selected features as $\mathcal{N} (0| u_{\ell_1 i}, SD)$  in BTuD based unsupervised FE that are identical  to 
$\mathcal{N} (u_{\ell_1 i}| 0, SD)$ in TD based unsupervised FE, the attributed $P$-val{\color{black}u}es are identical as well. Thus, both methods attribute the same $P$-values to selected and not selected features, respectively. This means that both methods are empirically identical (Fig. \ref{fig:exp}). 

One should  also notice that the equivalence between TD and BTuD based unsupervised FE stands only when TD and BTuD give the common $u_{\ell_1 i}$ by which we can select $i$s.
In Fig. \ref{fig:venn}, we were forced to treat $\ell_2=5,6$ as one group, since $u_{\ell_1 i}, \ell=5,6$ used to attribute $P$-values to $i$s are not identical between TD and BTuD.
Nevertheless,  $\ell_2=5,6$ as one group are coincident between TD and BTuD; it is biologically reasonable since  $\ell_2=5,6$ are associated with the distinct combination of two common tissues specificity (Fig. \ref{fig:tissue}).

\section{Conclusions}

In this paper, we have proposed a new implementation of Bayesian Tucker decomposition (BTuD) to perform unsupervised feature selection. 
BTuD was practically {\color{black}was practically implemented} by HOOI and we can attributed $P$-values to features with following Bayesian theory. BTuD based unsupervised FE works well {\color{black}for} various {\color{black}datasets}. BTuD is also expected to coincide with the previously proposed TD based unsupervised FE. 

\section{Acknowledgement}

This work was supported by JSPS KAKENHI 22K13979, 23K28150, 24K15168, 24K22309, 24H00247, 25K00986, 25H01470; JST; and PRESTO Grant numbers JPMJPR212A, JST CREST JPMJCR2431, and NEDO JPNP22100843-0.


\bibliography{sn-bibliography}

\end{document}